\documentclass[10pt,twocolumn]{IEEEtran}
\ifCLASSINFOpdf

\else

\fi

\usepackage{cite}
\usepackage{amsthm}
\usepackage{graphicx}
\usepackage{epstopdf}
\usepackage{amsmath}
\usepackage{amsfonts}
\usepackage{amssymb}
\usepackage{color}
\usepackage{xcolor}
\usepackage{multirow}
\usepackage{multicol}
\usepackage[utf8]{inputenc}
\usepackage[english]{babel}
\usepackage[algo2e]{algorithm2e}
\usepackage{algorithm}
\usepackage{hyperref}
\usepackage{nomencl}
\makenomenclature

\definecolor{orange}{RGB}{0,112,192}

\usepackage[textwidth=30mm]{todonotes}

\begin{document}
\title{Downlink Power Allocation in Massive MIMO via Deep Learning: Adversarial Attacks and Training 
}
\author{B.~R.~Manoj,~\IEEEmembership{Member,~IEEE,}
        Meysam~Sadeghi,~\IEEEmembership{Member,~IEEE,}
        and~Erik~G.~Larsson,~\IEEEmembership{Fellow,~IEEE}
\normalsize
\thanks{
This work was supported in part by Security-Link and the SSF SURPRISE project.

A part of this paper is published in the IEEE International Conference on Communications (ICC), 2021 \cite{Manoj_adversarial_attacks_2021}.

B. R. Manoj is with the Department of Electronics \& Electrical Engineering, Indian Institute of Technology Guwahati, Guwahati 781039, India (e-mail: {\tt  manojbr@iitg.ac.in}). 

Meysam Sadeghi was with Link\"{o}ping University, Link\"{o}ping, Sweden (e-mail: {\tt m.sadeghee@gmail.com}).

Erik G. Larsson is with the Department of Electrical Engineering (ISY), Link\"{o}ping University, Link\"{o}ping, Sweden (e-mail: {\tt erik.g.larsson@liu.se}).
}
}
\maketitle

\begin{abstract}
The successful emergence of deep learning (DL) in wireless system applications has raised concerns about new security-related challenges. One such security challenge is adversarial attacks. Although there has been much work demonstrating the susceptibility of DL-based classification tasks to adversarial attacks, regression-based problems in the context of a wireless system have not been studied so far from an attack perspective.  The aim of this paper is twofold: (i) we consider a regression problem in a wireless setting and show that adversarial attacks can break the DL-based approach and (ii) we analyze the effectiveness of adversarial training as a defensive technique in adversarial settings and show that the robustness of DL-based wireless system against attacks improves significantly. Specifically, the wireless application considered in this paper is the DL-based power allocation in the downlink of a multicell massive multi-input-multi-output system, where the goal of the attack is to yield an infeasible solution by the DL model. We extend the gradient-based adversarial attacks: fast gradient sign method (FGSM), momentum iterative FGSM, and projected gradient descent method to analyze the susceptibility of the considered wireless application with and without adversarial training. We analyze the deep neural network (DNN) models performance against these attacks, where the adversarial perturbations are crafted using both the white-box and black-box attacks. 
\end{abstract}


\section{Introduction}
Deep learning (DL) has become extremely popular due to its ability to efficiently learn end-to-end strategies from raw inputs and owing to the significant recent  increase in availability of computational power \cite{Schmidhuber_deep_learning_2015}. DL-based techniques have been applied widely in machine learning (ML) applications such as computer vision, natural language processing, medical diagnosis, and market trading.
The success of DL in different domains and the need for more complex models to characterize future wireless networks \cite{Huang_deep_learning_2020, Erpek_deep_learning_2020,Lian_RL_2020}, has stimulated a surge of interest to adopt DL in wireless communication applications, for example, power allocation in massive multi-input-multi-output (MIMO) \cite{Sanguinetti_deep_learning_2018}, channel estimation and signal detection \cite{Ye_power_of_2017}, channel decoding \cite{Liang_an_iterative_2018}, indoor user positioning \cite{Arnold_novel_massive_2019}, indoor activities sensing and classification using radio signals \cite{manoj2021sensing}, modulation classification \cite{Sadeghi_adversarial_attacks_2019, Kim_over_the_2020, Filipovic_adversarial_examples_2019, Flowers_evaluating_adversarial_2020, Bair_on_the_2019}, and the design of end-to-end communication systems \cite{Dorner_deep_learning_2018, Sadeghi_physical_adversarial_2019}.

It is important that DL models are reliable and robust to security threats before being deployed in the physical world, and this is  a particular concern when DL is applied to wireless communications.  In spite of promising performance, however, it has been found that DL-based systems can be vulnerable and exposed to new security related issues \cite{Hameed_the_best_2021, Erpek_deep_learning_2019, Sagduyu_adversarial_deep_2021, Shi_over_the_2020, Davaslioglu_trojan_attacks}. 
The different classes of security threat models that are of interest are inference attacks \cite{Shi_over_the_2020}, adversarial attacks \cite{Sadeghi_adversarial_attacks_2019}, and poisoning attacks \cite{Davaslioglu_trojan_attacks}. 

In this paper, our focus is only on adversarial attacks against models built with deep neural networks (DNNs). In adversarial attacks, a well-crafted adversarial perturbation is added to the input of the DNN that causes erroneous prediction (e.g., misclassification) \cite{Sadeghi_adversarial_attacks_2019,  Kim_over_the_2020, Filipovic_adversarial_examples_2019, Flowers_evaluating_adversarial_2020, Bair_on_the_2019, Sadeghi_physical_adversarial_2019}. The nature of the attack against DNN depends on the adversary's knowledge (that is, the attacker's information) and is categorized into white-box and black-box attacks. In white-box attacks, the adversary has access to all   information including the trained model and the model parameters \cite{Yuan_adversarial_examples_2019}. Although this type of attack may not be realistic, it helps to evaluate vulnerabilities and robustness against the models for the worst-case attack.  In black-box attacks, the adversary has limited information or no access to the trained model \cite{Yuan_adversarial_examples_2019}. 

In the literature, there has been a significant amount of work on adversarial attacks in DL-based wireless communication systems mainly focusing on classification based applications (e.g., radio signal classification \cite{Sadeghi_adversarial_attacks_2019,  Kim_over_the_2020, Filipovic_adversarial_examples_2019, Flowers_evaluating_adversarial_2020, Bair_on_the_2019}), while the regression-based applications (e.g., resource allocation \cite{Sanguinetti_deep_learning_2018}) are overlooked. In this paper, we take a first step towards filling this gap by investigating different types of adversarial attacks and then a defensive technique against the DNN model built for optimal power allocation  in the downlink of a massive MIMO network. To enable reproducibility of the results presented in this paper, we have considered the openly available dataset and the DNN models of \cite{Sanguinetti_deep_learning_2018}.

Recently, in the context of wireless application, only \cite{Kim_adversarial_attacks_2021} used a regression-based DNN model to allocate the power in the downlink scenario for orthogonal subcarriers and then showed that by launching adversarial attacks, that is by manipulating the inputs to the DNN, it is possible to reduce the minimum rate among all users. Different from \cite{Kim_adversarial_attacks_2021}, in this paper, for our considered regression problem in a wireless network setting, the focus is to investigate different adversarial attacks to generate adversarial examples in terms of attacker success rate. Then, we analyze the defense technique through adversarial training for robustness against these attacks. More specifically, our main contributions in this paper are as follows:
\begin{itemize}
\item[(a)] The most commonly used adversarial attacks in computer vision and image classification applications are the  fast gradient sign method (FGSM) \cite{Sadeghi_adversarial_attacks_2019}, the momentum iterative FGSM \cite{Dong_boosting_adversarial_2018}, and the projected gradient descent method (PGDM) \cite{Madry_towards_deep_2018}. We extend these attacks to generate adversarial examples for the regression problem in the context of a massive MIMO system. In particular, we consider the problem of optimal power allocation using the product signal-to-interference-plus-noise ratio (SINR) maximization strategy in the downlink scenario using maximal-ratio (MR) and multicell-minimum mean square error (M-MMSE) precoding schemes. 

\item[(b)] We investigate both white-box and black-box attacks against DNN models. The aim of the adversarial attack is to create an infeasible solution by the DNN. We benchmark our proposed adversarial attacks and show that PGDM attack is the most successful  in fooling the DNN. We further demonstrate by assuming that the adversary having access to the input of the DNN model, i.e., we perform a white-box attack, the adversary can cause up to 85\% of infeasible solutions to be generated by the proposed DNN models of \cite{Sanguinetti_deep_learning_2018}. 

\item[(c)] Considering a worst-case scenario, we show that adversarial training   is not only a very efficient method to improve the robustness of DNN models against adversarial attacks, but also to boost the robustness of networks against natural unwanted disturbances such as  random noise.

\item[(d)] Finally, we use the property of transferability to characterize black-box attacks and analyze the effectiveness of this property for MR and M-MMSE precoding schemes, i.e., we study the adversary success rate of black-box attacks. 
\end{itemize}

\section{System Model}
We consider a multicell massive MIMO system having $L$ cells, where each cell consists of a base station (BS) with $M$ antennas and $K$ single-antenna user equipments (UEs). 
The channel between UE $k$ in cell $l$ and the BS in cell $j$ is denoted by 
${\bf{h}}_{lk}^j \in \mathbb{C}^{M \times 1}$ and is modeled as
\begin{equation}
{\bf{h}}_{lk}^j \sim {\mathcal{N}_{\mathbb{C}}}(0, {\bf{R}}_{lk}^j)\, , 
\end{equation}
where ${\bf{R}}_{lk}^j \in \mathbb{C}^{M \times M}$ is the spatial correlation matrix. The distance from BS in cell $j$ to UE $k$ in cell $l$
is denoted by $d_{lk}^j$,  where $d_{lk}^j = ||{\bf{z}}_{lk}^j||$ and 
${\bf{z}}_{lk}^j \in \mathbb{R}^{2 \times 1}$ is the location of UE in the Euclidean space. That is, $z^j_{lk}$ is location vector in the local coordinate system with base station $j$ as origin. It is assumed that during the uplink pilot transmission phase, $\tau_p = K$ pilots are utilized \cite{Sanguinetti_deep_learning_2018}.

The signal transmitted in the downlink by the BS in cell $j$ is given as ${\bf{x}}_j = \sum_{k=1}^{K} {\bf{w}}_{jk} \,s_{jk}$, where $s_{jk} \sim {\cal{N}}(0, \rho_{jk})$ denotes the downlink data signal for user $k$ in the cell, ${\bf{w}}_{jk} \in \mathbb{C}^{M \times 1}$ denotes the associated precoding vector that determines the transmission beamforming with $||{\bf{w}}_{jk}||^2 = 1$ being satisfied, and $\rho_{jk}$ is the signal transmission power. 
We consider the MR and M-MMSE combining as the precoding methods which have been adopted by the authors in \cite{Sanguinetti_deep_learning_2018}. The downlink SINR of UE $k$ in cell $j$ is given by
\begin{equation}
\gamma_{jk} = \frac{\rho_{jk} \; a_{jk}}{\sum_{l=1}^{L} \sum_{i=1}^{K} \rho_{li} \; b_{lijk} + \sigma^2} \, , 
\label{SINR_eq}
\end{equation}
where $\sigma^2$ is the variance of additive white Gaussian noise (AWGN). The average channel and the interference gains [ch.~7]\cite{bjornson2017massive}, are respectively given as
\begin{equation}
a_{jk} = |\mathbb{E}\{{\bf{w}}_{jk}^{\text{H}} \, {\bf{h}}_{jk}^j \}|^2\, ,
\label{avg_ch_eq}
\end{equation}
and
\begin{equation}
b_{lijk} = 
\begin{cases}
     \mathbb{E}\{|{\bf{w}}_{li}^{\text{H}} \, {\bf{h}}_{jk}^l|^2\}, \qquad \qquad \qquad \quad \quad \, \,\, \text{if}\,\, (l,i) \neq (j,k) \\
    \mathbb{E}\{|{\bf{w}}_{jk}^{\text{H}} \, {\bf{h}}_{jk}^j|^2\} - |\mathbb{E}\{{\bf{w}}_{jk}^{\text{H}} \, {\bf{h}}_{jk}^j\}|^2, \quad \text{if}\, \, (l,i) = (j,k)
\end{cases}
\label{avg_interf_eq}
\end{equation}
where $\mathbb{E}\{\cdot\}$ is the expectation operator and $(\cdot)^{\text{H}}$ denotes the Hermitian transpose. The achievable spectral efficiency (SE) $\mathrm{log}_2(1+\gamma_{jk})$, with $\gamma_{jk}$ obtained from \eqref{SINR_eq}, is a rigorous lower bound on ergodic Shannon capacity \cite{mMIMO2016fundamentals}. This bound relies on the assumption that the receiver knows the statistical expectation of the effective channel (which is a deterministic constant) \cite{mMIMO2016fundamentals}. 
The effective SINR used here is well established in the literature, information-theoretically rigorous and derived from first principles for example in  \cite{bjornson2017massive}. 

The work in this paper aims at analyzing various adversarial attacks on the DNN models for the regression-based wireless application, and then study the potential of adversarial training to improve the robustness of these  models. Thus, in this regard, to investigate the findings in detail, we consider the product SINR maximization-based optimal power allocation strategy, which is formulated as 
\begin{eqnarray}\label{max-prod_eq}
&& \underset{\rho_{jk}: \, \forall j, k}{\max}  \,\, 				\prod_{j=1}^{L} \prod_{k=1}^{K} \gamma_{jk}\,,    
\\&&
\text{s.t.}  \, \, \sum_{k=1}^{K} \rho_{jk} \leq P_{\mathrm{max}}\, , \, j = 1,\ldots, L\,, \nonumber 
\end{eqnarray}
where $P_{\mathrm{max}}$ is the total downlink transmit power.

\section{Dataset and the deep neural network models}
To investigate the security related issues of a DNN-based wireless network in a regression setting, we  use the dataset that is publicly available for DL-based power allocation in the downlink of a multicell massive MIMO system and its associated DNN architectures \cite{Sanguinetti_deep_learning_2018}. In \cite{Sanguinetti_deep_learning_2018}, the goal of using DL data driven-based approach is to allocate powers optimally to the UEs by using the max-product power allocation policy. 

We denote the dataset as $\{{\bf{u}}(n),{\bf{v}}(n)\}_{n=1}^{N_1}$ with ${\bf{u}}(n)$ as the input data sample and  ${\bf{v}}(n)$ as the output data sample, and ${N}_1$ is the size of the training dataset. The DNN model for the considered regression-based application is defined as $f(.;{\boldsymbol{\theta}})$, where  ${\bf{u}}(n)$ is the input to the model and $f({\bf{u}}(n)) = {\bf{v}}(n)$ is the predicted output, and ${\boldsymbol{\theta}}$ is the hyperparameters of the model $f(\cdot)$. The loss function of $f(\cdot)$ is denoted by ${\cal{L}}({\boldsymbol{\theta}},{\bf{u}}(n),{\bf{v}}(n))$. For a given multicell massive MIMO system, the aim of the DNN model is to learn the mapping between the positions of UEs denoted as  
${\bf{x}}(n) {\in \mathbb{R}}^{2KL \times 1}$ and the solution of optimal power coefficients in the cell $j$ denoted as ${\boldsymbol{\rho}}_j(n) = [\rho_{j1}, \hdots, \rho_{jK}] {\in \mathbb{R}}^{K \times 1}$, $j = 1, \ldots, L$. This implies that the DNN learns the mapping relation between the $2KL$ UE positions and the power allocation solution ${\boldsymbol{\rho}}_j(n)$ to (\ref{max-prod_eq}) for the individual cell $j$. 
The optimal solution is obtained by solving (\ref{max-prod_eq}) using (\ref{avg_ch_eq}) and (\ref{avg_interf_eq}) through  conventional optimization methods \cite{bjornson2017massive}.  The optimization problem in (\ref{max-prod_eq}) can be solved by geometric programming, requiring a polynomial or quasi-polynomial complexity \cite{Sanguinetti_deep_learning_2018, bjornson2017massive}.

The associated DNN architecture considered in \cite{Sanguinetti_deep_learning_2018} is the feedforward neural network (NN) with fully connected layers comprised of an input layer of $2KL$ dimension, $N$ hidden layers, and an output layer of $K+1$ dimension. The output layer produces an estimate 
${\boldsymbol{\hat{\rho}}}_j = [\hat{\rho}_{j1}, \hdots, \hat{\rho}_{jK}]{\in \mathbb{R}}^{K \times 1}$ of the optimal solution ${\boldsymbol{\rho}}_j = [\rho_{j1}, \hdots, \rho_{jK}] {\in \mathbb{R}}^{K \times 1}$. The dimension of the output layer is $K+1$ instead of $K$, since the authors in \cite{Sanguinetti_deep_learning_2018} make use of the NN  learns the constraint $\sum_{k=1}^{K} \rho_{jk} \leq P_{\mathrm{max}}$ to improve the accuracy of ${\boldsymbol{\hat{\rho}}}_j$ estimation. For the considered application, using DL-based approach has the following fundamental benefits: 
\begin{itemize}
    \item Solving the problem at hand requires a quasi-polynomial algorithm \cite{Sanguinetti_deep_learning_2018}. However, even a polynomial complexity can be too much when the solution must be obtained in real-time; that is, fast enough to be deployed in the system before the UEs' positions change and the power allocation problem needs to be solved again \cite{Sanguinetti_deep_learning_2018}. While using DNN, the problem can be solved online. That is, after configuring the parameters such as weight and bias, the DNN can estimate ${\hat{\boldsymbol{\rho}}}_j$ for UEs that are not part of the training set. This implies that every time UEs positions change in the network, their updated  ${\hat{\boldsymbol{\rho}}}_j$ can be obtained by feeding the new positions to the DNN, without solving \eqref{max-prod_eq}.
    
    \item The DNN approach for power allocation just requires the location information of UEs for computing the powers at any given cell. This is in sharp contrast to the optimization approach, as the optimization approach requires the UEs channel state information, which, in turn, enforces further complexities, like estimating the channel state information for UEs.
\end{itemize}

The network parameters that are considered for the power allocation in  \cite{Sanguinetti_deep_learning_2018} are: number of cells $L=4$, number of UEs in each cell $K=5$, number of antennas at the BS $M = 100$, maximum transmit power in the downlink $P_{\mathrm{max}} = 500$mW, $\sigma^2 = -94$dBm, and the communication bandwidth is $20$ MHz. Each cell has a square area of $250$m $\times$ $250$m and is deployed on a square grid layout of $2\times2$ cells, where each square has a BS in the center. The network uses a wrap-around topology.

During the training phase of the considered system, the DNN is trained with $N_1$ samples taking input as the UE positions ${\bf{x}}(n)$ and the output is the optimal power values of cell $j$ as ${\boldsymbol{\rho}}_j(n)$, where $A = \{n:n = 1, \ldots, N_1\}$ and $j = 1, \ldots, L$.
The dimensions of input and output samples of DNN are $2KL = 40$ and $K+1 = 6$, respectively. 
During the test phase, we denote the input to DNN as ${\bf{x}}_\text{t}(m)$ and the predicted output power values of the DNN $f({\bf{x}}_\text{t}(m))$ as 
${\boldsymbol{\hat{\rho}}}_j(m) = [\hat{\rho}_{j1}, \hdots, \hat{\rho}_{jK}]$, where $B =\{m:m = N_1+1, \ldots, N_1+N_2\}$ with $A \cap B = \emptyset$ (i.e., the test and training samples are different) and $N_2$ is the size of the test dataset. 

The DNN architecture employed for solving the max-product SINR power allocation problem with M-MMSE and MR combining techniques \cite{Sanguinetti_deep_learning_2018} is shown in Table \ref{tab:table1}, consisting of $6,981$ trainable parameters. In \cite{Sanguinetti_deep_learning_2018}, for both M-MMSE and MR schemes, the performance of the data-driven approach (i.e., estimation of power allocations through DNN) is compared with that of the model-driven   approach (i.e., computation of power allocations through conventional optimization methods \cite{bjornson2017massive}) by evaluating the average MSE. Therefore, to further reduce the average MSE as compared to the model in Table \ref{tab:table1}, a more complex DNN is employed in \cite{Sanguinetti_deep_learning_2018} as depicted in Table \ref{tab:table2}, consisting of $202,373$ trainable parameters. For convenience, we refer to the DNN models in Table \ref{tab:table1} and Table \ref{tab:table2} as $M_1$ and $M_2$, respectively.  
\begin{table}[t!]
  \begin{center}
    \caption{Architecture $M_1$: DNN model with trainable parameters of $6,981$.}
    \label{tab:table1} \scalebox{0.9}{
    \begin{tabular}{|c|c|c|c|}  
        & Size & Parameters & Activation function \\
      \hline
      Input & 40 & - & - \\  \hline
      Layer 1 (Dense) & 64 & 2624 & elu \\ \hline
      Layer 2 (Dense) & 32 & 2080 & elu \\ \hline
      Layer 3 (Dense) & 32 & 1056 & elu \\ \hline
      Layer 4 (Dense) & 32 & 1056 & elu \\ \hline
      Layer 5 (Dense) & 5  & 165  & elu \\ \hline
      Layer 6 (Dense) & 6  & 36   & linear \\ \hline
    \end{tabular} }
  \end{center}
\end{table}

\begin{table}[t!]
  \begin{center}
   \vspace{-0.25em}
   \caption{Architecture $M_2$: DNN model with trainable parameters of $202,373$.}
    \label{tab:table2} \scalebox{0.9}{
    \begin{tabular}{|c|c|c|c|}  
        & Size & Parameters & Activation function \\
      \hline
      Input & 40 & - & - \\  \hline
      Layer 1 (Dense) & 512 & 20992  & elu \\ \hline
      Layer 2 (Dense) & 256 & 131328 & elu \\ \hline
      Layer 3 (Dense) & 128 & 32896  & elu \\ \hline
      Layer 4 (Dense) & 128 & 16512  & elu \\ \hline
      Layer 5 (Dense) & 5   & 645    & elu \\ \hline
      Layer 6 (Dense) & 6   & 36     & linear \\ \hline
    \end{tabular}}
  \end{center}
\end{table}

\section{Adversarial attacks} 
\begin{table}[!t]
	 \caption{Notations.} 
	\begin{center}
		\renewcommand{\arraystretch}{1.45}
		\scalebox{0.86}{
		\begin{tabular}{|c|c|}
			\hline
	 $f(.;\boldsymbol{\theta})$ & DNN model for the regression-based task \\ \hline
	 $\boldsymbol{\theta}$ & Hyperparameters of model $f$ \\ \hline
	${\cal{L}}(\cdot)$ & Loss function of model $f$ \\ \hline
	${\bf{x}}_\text{t}$ & Original input data (clean UE positions) \\ \hline
	${\bf{x}}_\textrm{adv}$ &  Adversarial perturbed data (perturbed UE positions) \\ \hline
	${\boldsymbol{\rho}}_j$ & True output power vector of cell $j$ (optimal solution) \\ \hline 
	${\boldsymbol{\hat{\rho}}}_j$ & Predicted output power vector by $f$ in the cell $j$ for the input  ${\bf{x}}_\text{t}$ \\ \hline
	${\boldsymbol{\hat{\rho}}}^\text{adv}_j$ & Predicted output power vector by $f$ in the cell $j$ for the input  ${\bf{x}}_\textrm{adv}$ \\ \hline
    ${\hat{\rho}}_{jk}$ & Power coefficient of UE $k$ in cell $j$ \\ \hline
    ${\boldsymbol{\eta}} = {\bf{x}}_\textrm{adv} -{\bf{x}}_\text{t}$ & Adversarial perturbation \\ \hline
    $||\cdot||_p$ &  $L_p$-norm \\ \hline 
    $\epsilon$ &  Perturbation magnitude \\ \hline 
    $d_{\epsilon} \leq \sqrt{2}\, \epsilon$ & Distance perturbation \\ \hline 
    $P_{\mathrm{max}}$ & Downlink transmission power \\ \hline 
    $\nabla_{{\bf{a}}} {{\bf{b}}}$ & Gradient of ${{\bf{b}}}$ with respect to ${{\bf{a}}}$ \\
			\hline
		\end{tabular}
 }
	\end{center}
	\label{table_notations} 
\end{table}
In this section, we present discussion on common adversarial attacks in classification-based applications and extend them to a regression-based problem in the context of power control for massive MIMO wireless systems. Specifically, we propose the generation of adversarial examples assuming that the adversary has access to the DNN and the UEs positions. This, in turn, helps to obtain the adversary success rate in terms of infeasible output power (i.e., sum of UE powers in each cell $j$ is larger than ${P}_\mathrm{max}$) of DNN models for the worst-case assumption, enabling a study of the effectiveness of adversarial training as a defense mechanism. In each cell, we analyze that given $N_2$ adversarial examples as input to the DNN, for how many of theses examples the DNN provide infeasible power solutions as output. We perform this across all the cells, therefore, we get the total number of examples that provide infeasible output. Finally, to obtain the adversary success rate, we divide these total number of infeasible examples by $LN_2$. Therefore, for the generation of adversarial examples, we first present the algorithms for input-specific 
white-box attacks and then extend these attacks to the black-box case using transferability. For convenience, we provide the important notations in Table \ref{table_notations}.  

We consider the model $f({\bf{x}}_\text{t})$: ${\bf{x}}_\text{t} \in {\cal{X}} \rightarrow {\boldsymbol{\hat{\rho}}}_j \in {\cal{P}}$ with ${\bf{x}}_\text{t}$ as the input and the predicted output in the cell $j$ as ${\boldsymbol{\hat{\rho}}}_j$, where ${\cal{X}} \subset	{\mathbb{R}}^{\lambda}$ and ${\cal{P}} \subset	{\mathbb{R}}^{\nu}$ having the dimensions of input and output samples as $\lambda$ and $\nu$, respectively. 
The goal of the adversary is to generate the adversarial perturbed samples denoted by ${\bf{x}}_\textrm{adv}$ in the neighborhood region of the original clean samples ${\bf{x}}_\text{t}$ under a given constraint causing the predicted output of the model erroneous. This can be written as 
\begin{eqnarray}
&& {\bf{x}}_\textrm{adv} = {\bf{x}}_\text{t} + {\boldsymbol{\eta}} \\ &&
{\text{such that}}\,\, f({\bf{x}}_\text{t}) = {\boldsymbol{\hat{\rho}}}_j \,, \,\, f({\bf{x}}_\text{adv}) = {\boldsymbol{\hat{\rho}}}_j^\text{adv}\,, \nonumber \\ &&
\qquad \qquad \sum_{k=1}^{K} {\hat{\rho}}_{jk} \leq  P_\text{max}\,, \, \, \sum_{k=1}^{K} {\hat{\rho}}^\text{adv}_{jk} > P_\text{max} \nonumber \\ && 
\qquad \qquad {\bf{x}}_\text{t}\,,  {\bf{x}}_\text{adv} \in {\cal{X}}\,, 
\nonumber \\ && 
\qquad \qquad \mathrm{and}\,\, ||{\boldsymbol{\eta}}||_\infty\, \leq \epsilon\,, \nonumber 
\end{eqnarray}
where ${\boldsymbol{\hat{\rho}}}_j = [\hat{\rho}_{j1}, \hdots, \hat{\rho}_{jK}]$ and ${\boldsymbol{\hat{\rho}}}^\text{adv}_j = [\hat{\rho}^\text{adv}_{j1}, \hdots, \hat{\rho}^\text{adv}_{jK}]$ are the predicted outputs by the model $f(\cdot)$ for the inputs ${\bf{x}}_\text{t}$ and ${\bf{x}}_\text{adv}$, respectively, 
${\boldsymbol{\eta}}$ is a perturbation added to ${\bf{x}}_\text{t}$ with $\epsilon$ as the perturbation magnitude and $||{\boldsymbol{\eta}}||_\infty$ is the $L_\infty$-norm of the adversarial perturbation.

\subsection{Attack model}
\begin{figure}[t]
	\centering
	\includegraphics[width=3.65in,height=1.3in]{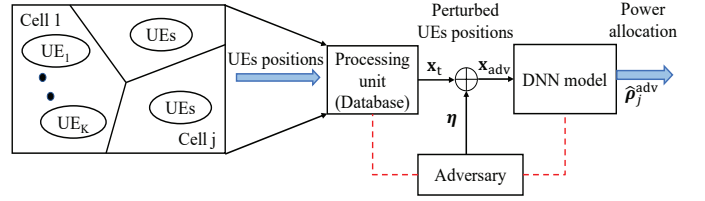}
	\caption{White-box attack model for DL-based power allocation in a multicell massive MIMO system.}
	\label{threat_model}
\end{figure}
We envisage the attack model for the considered DL-based power allocation application as depicted in Fig. \ref{threat_model}.
In each cell $j$, after obtaining the UEs positions through global positioning system (GPS), all the UEs transmit their locations to the central processing unit (CPU). The locations of UEs across all cells (referred to as ${\bf{x}}_\text{t}$) are then fed to the DNN model to predict the optimal power control coefficients of UEs of each cell. In this paper, we employ the  worst-case assumptions that the adversary referred to as the hostile entity (i) has access to the input of the DNN and (ii) knows the locations of the UEs. These assumptions enable us to analyze the effectiveness of adversarial training in a worst-case attack scenario, thus, obtaining a lower-bound on the robust performance against adversarial attacks. 

The adversary aims to compute the adversarial perturbation such that the loss function is maximized so that the DL-based wireless system breaks down. The adversarial perturbation can be computed in the direction of the gradient to maximize the loss function under the constraint  $||{\boldsymbol{\eta}}||_\infty \leq \epsilon$, where ${\boldsymbol{\eta}}$ is a very small quantity which is in the order of centimetres (cms) as compared to that of the actual UE position values. The positions of the users obtained through GPS information has a typical error range within a couple of meters. Therefore, we limit the malicious adversarial perturbations to be very small within this error range so that it cannot be diagnosed with any detection algorithm at the BS since this will be well below the tolerance range of the GPS. Thus, in this paper, we confine ourselves to $\epsilon<0.5$ in order to keep the attacks practical and realizable. If $\epsilon$ is large, then it might be possible that the BS could detect these malicious adversarial perturbations by utilizing a coarse grained location estimate independently of UE reports.

The physical interpretation of adversarial perturbation is to modify the location of UEs position from the actual value to a new value by an amount of $d_{\epsilon}$ in the direction such that the loss function is increased, where $d_{\epsilon}$ denotes the distance perturbation. The focus of this paper is to first study from the attacker's point of view the success rate of the attacks and then to devise and analyze a defense  mechanism through adversarial training to create the robust networks designed for wireless application.

\subsection{Loss function} 
The generation of adversarial examples using various attacks focus on maximizing some loss function caused by $||{\boldsymbol{\eta}}||_\infty \leq \epsilon$. The loss function is designed such that the technique of crafting an adversarial example exists allowing the adversary to choose an attack mechanism to fool the model with a high probability and low computational cost. We design such a loss function to be maximized as 
\begin{equation}
{\cal{L}}({\boldsymbol{\theta}},\cdot) = \sum_{k=1}^{K} {\hat{\rho}}_{jk}\, ,
\label{loss_function}
\end{equation}
such that the DNN model predicts the output that is infeasible power solution and the DL-based wireless system is collapsed, where ${\hat{\rho}}_{jk}$ is the predicted power by the model $f(\cdot)$ for UE $k$ in cell $j$ for an clean input sample  ${\bf{x}}_\text{t}$. The infeasible power solution is defined as $\sum_{k=1}^{K} {\hat{\rho}}_{jk} > P_{\textrm{max}}$, because  this inequality can never be satisfied. 
In the context of downlink power allocation to UEs in a multicell massive MIMO system by using the DL-based approach, given clean input ${\bf{x}}_\text{t}$ and the trained model $f(\cdot)$, the generation of ${\bf{x}}_\text{adv}$ can be achieved by formulating the following optimization problem as 
\begin{eqnarray}
\label{adv_optmiz}
 && \hspace{-1em} \underset{{\bf{x}}_{\text{adv}}}{\text{argmin}} \left\lVert {\bf{x}}_{\text{adv}}- {\bf{x}}_\text{t}\right\rVert_\infty  \\ &&
\hspace{-1em} \text{s.t.}  \quad f({\bf{x}}_\text{adv}) = {\boldsymbol{\hat{\rho}}}^\text{adv}_j\,,\,\,  \sum_{k=1}^{K} {\hat{\rho}}^\text{adv}_{jk} > P_\text{max}\,, \,\, {\bf{x}}_{\text{adv}} \in {\cal{X}}\, , \nonumber  \\ && 
\hspace{-1em} \text{and} \quad \left\lVert {\bf{x}}_{\text{adv}}- {\bf{x}}_\text{t}\right\rVert_\infty \leq \epsilon\,. \nonumber 
\end{eqnarray}
For UEs positions ${\bf{x}}_\text{t}$, $f({\bf{x}}_\text{t})$ predicts power allocation ${\boldsymbol{\hat{\rho}}}_j = [\hat{\rho}_{j1}, \hdots, \hat{\rho}_{jK}]$ for all the UEs in cell $j$. The sum of predicted powers by the model $f(\cdot)$ in each cell $j$ is constrained as $\sum_{k=1}^{K} {\hat{\rho}}_{jk} \leq P_{\mathrm{max}}$, which we refer to as a feasible solution of the model. The perturbed UEs positions ${\bf{x}}_\text{adv} = {\bf{x}}_\text{t} + {\boldsymbol{\eta}}$, is also within the area of cell $j$, thus, we expect  $f({\bf{x}}_{\text{adv}})$ to predict the power coefficients in the same order as $f({\bf{x}}_\text{t})$ as long as the $||{\boldsymbol{\eta}}||_\infty \leq \epsilon$ is satisfied. However, this does not hold, with adversarial perturbation, 
the output of the model  
$f({\bf{x}}_{\text{adv}})$ predicts adversarial power as ${\boldsymbol{\hat{\rho}}}^\text{adv}_j = [\hat{\rho}^\text{adv}_{j1}, \hdots, \hat{\rho}^\text{adv}_{jK}]$ which could provide infeasible solution, i.e.,  $\sum_{k=1}^{K} {\hat{\rho}}^\text{adv}_{jk} > P_{\mathrm{max}}$ as shown in (\ref{adv_optmiz}) if we could maximize some loss function as given in (\ref{loss_function}) such a way that the system collapses. This implies that the perturbation of UE position values are not random noise, rather they are displaced in   well-crafted directions. 
The generation of adversarial examples by using the optimization problem in (\ref{adv_optmiz}) is in practice difficult to solve due to its intractability. Therefore, to generate the adversarial examples efficiently in terms of computational cost and time, there are many sub-optimal methods in the literature \cite{Sadeghi_adversarial_attacks_2019, Yuan_adversarial_examples_2019, Dong_boosting_adversarial_2018}.

In this paper, the class of gradient-based methods  \cite{Yuan_adversarial_examples_2019, Dong_boosting_adversarial_2018, Madry_towards_deep_2018} are considered using the  $L_{\infty}$-norm of the perturbation  constraint. Specifically, the adversarial perturbation constraint is given by $||{\boldsymbol{\eta}}||_{\infty} \leq \epsilon$, implying that $L_{\infty}$-norm is upper bounded  by the perturbation magnitude $\epsilon$. The motivation behind choosing $L_{\infty}$-norm is that, the adversarial perturbation to the clean UE positions provides the maximum distance perturbation ($d_\epsilon$) from the actual positions. The geometrical interpretation for $L_{\infty}$-norm bound is that the perturbations of UE positions correspond to the maximum displacement in the direction of the gradient of loss function such that the loss in the steepest direction is increased. The distance perturbation and  the perturbation magnitude are related as $d_{\epsilon} \leq \sqrt{2}\, \epsilon$.

\subsection{Fast gradient sign method (FGSM)}
The FGSM was proposed by Goodfellow \cite{Goodfellow_explaining_and_2015} to determine the adversarial examples in a computationally efficient manner.
In this method, adversarial samples ${\bf{x}}_{\text{adv}}$
are generated by maximizing the loss function ${\cal{L}}({\boldsymbol{\theta}},f({\bf{x}}_\text{t}))$ in the $L_\infty$-neighbor of the clean samples. FGSM is a typical one-step gradient update attack, where the perturbation is determined along the direction of the gradient of ${\cal{L}}({\boldsymbol{\theta}},f({\bf{x}}_\text{t}))$ that is expressed as  
\begin{eqnarray}
{\boldsymbol{\eta}} = \epsilon \, {\text{sign}}\left(\nabla_{{\bf{x}}_\text{t}} {\cal{L}}({\boldsymbol{\theta}},f({\bf{x}}_\text{t})) \right) \, ,
\label{fgsm_perturb}
\end{eqnarray}
where $\nabla_{{\bf{x}}_\text{t}} {\cal{L}}({\boldsymbol{\theta}},f({\bf{x}}_\text{t}))$ is the gradient of ${\cal{L}}({\boldsymbol{\theta}},f({\bf{x}}_\text{t}))$ which can be computed using backpropagation.
The adversarial example is then generated as ${\bf{x}}_\text{adv} = {\bf{x}}_{\text{t}} + {\boldsymbol{\eta}}$.
We propose a cell-based FGSM attack against downlink power allocation in a massive MIMO system, as shown in Algorithm~$1$. In a cell-based attack, the adversary attacks each cell by maximizing the loss function in the respective cell of the wireless network. 
In Algorithm~$1$, we denote ${\bf{x}}_\text{t} = [{\bf{x}}_\text{t}(1), \hdots, {\bf{x}}_\text{t}(m)]$, ${\bf{x}}_j^\text{adv} = [{\bf{x}}_j^\text{adv}(1), \hdots, {\bf{x}}_j^\text{adv}(m)]$, and ${\boldsymbol{\hat{\rho}}}_{j}^\text{adv}(m) = [{\hat{\rho}}_{j1}^\text{adv}, \hdots, {\hat{\rho}}_{jK}^\text{adv}]$ 
as the clean input test samples,  adversarial generated examples in cell $j$, and the predicted adversarial output power allocation of the model $f(\cdot)$ with adversarial perturbed input for the sample $m$ in cell $j$, respectively, where $m = 1, \hdots, N_\text{2}$ and $j = 1, \hdots, L$.
\begin{algorithm}[!t]
	\label{alg_fgsm}
	\SetAlgoLined
	\textbf{Input:} $f(.;{\boldsymbol{\theta}})$,  ${\bf{x}}_\text{t}(m)$, $\epsilon = \frac{d_{\epsilon}}{\sqrt 2}$, $L$, $P_{\mathrm{max}}$, $N_\text{2}$ \\
 	\textbf{Output:} ${\bf{x}}_j^\text{adv}(m)$, ${\boldsymbol{\hat{\rho}}}_{j}^\text{adv}(m) = [{\hat{\rho}}_{j1}^\text{adv}, \hdots, {\hat{\rho}}_{jK}^\text{adv}]$ \\
 	\textbf{Initialize:} $\mathrm{ix\_list} =$ [\,]
	
	\For {$j$ $\mathrm{in}$ range($L$)}
	{
	 \For {$m$ $\mathrm{in}$ range($N_{\mathrm{2}}$)}
	 {
	 ${\boldsymbol{\hat{\rho}}}_j(m) = [\hat{\rho}_{j1},\hdots,\hat{\rho}_{jK}] \leftarrow f({\bf{x}}_\text{t}(m))$\\	 
	 ${\cal{L}}({\boldsymbol{\theta}}, f({\bf{x}}_\text{t}(m))) = \sum_{k=1}^{K} {\hat{\rho}}_{jk}$ \\
     ${\boldsymbol{\eta}} = \epsilon \, {\text{sign}}\left(\nabla_{{\bf{x}}_\text{t}(m)} {\cal{L}}({\boldsymbol{\theta}}, f({\bf{x}}_\text{t}(m))) \right)$\\
     ${\bf{x}}_j^\text{adv}(m) = {\bf{x}}_\text{t}(m) + {\boldsymbol{\eta}}$\\
     ${\boldsymbol{\hat{\rho}}}_{j}^\text{adv}(m) = [{\hat{\rho}}_{j1}^\text{adv}, \hdots, {\hat{\rho}}_{jK}^\text{adv}] \leftarrow f({\bf{x}}_j^\text{adv}(m))$\\
\eIf{$\sum_{k=1}^{K} {\hat{\rho}}_{jk}^\mathrm{adv} >  P_{\mathrm{max}}$}
        {
			${\boldsymbol{\hat{\rho}}}_{j}^\text{adv}(m) \leftarrow$ infeasible power \\
			$\mathrm{ix\_list}.\mathrm{append}(m)$
		}
		{
		${\boldsymbol{\hat{\rho}}}_{j}^\text{adv}(m) \leftarrow$ feasible power	 
		}
	 }
	 ${\bf{x}}_j^\text{adv}(\mathrm{ix\_list}) $\\
	 ${\boldsymbol{\hat{\rho}}}_{j}^\text{adv}(\mathrm{ix\_list})$ 
	}
  \caption{FGSM attack}
\end{algorithm}

\subsection{Projected gradient descent method (PGDM)}
An extension of FGSM is to apply it with multiple iterations to perform a finer iterative optimizer. The PGDM attack is a more powerful attack which is also called   a multi-step iterative variant of FGSM attack \cite{Madry_towards_deep_2018}. In each iteration, the PGDM performs FGSM with a small step size $\alpha$ and projects the intermediate result of adversarial sample onto the $L_\infty$-norm around the clean sample to ensure that it is within the $\epsilon$-neighborhood of the clean sample ${\bf{x}}_\text{t}$ \cite{Madry_towards_deep_2018, maini_adversarial_robustness_2020}. The $q$-th step of the iteration in the  PGDM attack is given as follows
\begin{eqnarray}
&& \hspace{-1.5em} {\bf{x}}_{{\text{t}},0} = {\bf{x}}_{\text{t}}  \nonumber \\
&& \hspace{-1.5em} {\bf{x}}_{\text{t},q+1} = {\text{clip}}_{[{\bf{x}}_{\text{t}}, \epsilon]} \{{\bf{x}}_{\text{t},q} + \alpha \, \text{sign} \left(\nabla_{{\bf{x}}_{\text{t},q}} {\cal{L}}({\boldsymbol{\theta}},f({\bf{x}}_{\text{t},q}))\right) \}, \nonumber \\
\label{pgd_update}
\end{eqnarray}
where the small step size is denoted as $\alpha$ and  ${\text{clip}}_{[{\bf{x}}_{\text{t}}, \epsilon]}\{{\bf{x}}_{{\text{t}},q}\}$ is the element-wise clipping of ${\bf{x}}_{{\text{t}}, q}$ to lie in the range $[{\bf{x}}_{\text{t}} - \epsilon, {\bf{x}}_{\text{t}} + \epsilon]$ such that the intermediate result will be in the $L_{\infty}$ $\epsilon$-neighborhood of ${\bf{x}}_\text{t}$. The adversarial sample is determined after $Q$ iterations as $ {\bf{x}}_{\text{adv}} = {\bf{x}}_{\text{t},Q}$. The values of $Q$ and $\alpha$ are selected such that it is sufficient for the adversarial sample reaches the edge of the $L_{\infty}$ $\epsilon$-norm while limiting the computational cost to be minimum. We propose a cell-based PGDM attack against downlink power allocation, as shown in Algorithm~$2$. 
\begin{algorithm}[!t]
	\label{alg_pgd}
	\SetAlgoLined
	\textbf{Input:} $f(.;{\boldsymbol{\theta}})$,  ${\bf{x}}_\text{t}(m)$, $\epsilon = \frac{d_{\epsilon}}{\sqrt 2}$, $L$, $P_{\mathrm{max}}$, $\alpha$, $Q$, $N_\text{2}$ \\
 	\textbf{Output:} ${\bf{x}}_j^\text{adv}(m)$, ${\boldsymbol{\hat{\rho}}}_{j}^\text{adv}(m) = [{\hat{\rho}}_{j1}^\text{adv}, \hdots, {\hat{\rho}}_{jK}^\text{adv}]$\\
 	\textbf{Initialize:} $\mathrm{ix\_list} =$ [\,]
	
	\For {$j$ $\mathrm{in}$ range($L$)}
	{
	 \For {$m$ $\mathrm{in}$ range($N_{\mathrm{2}}$)}
	 {		 
		 \For {$q$ $\mathrm{in}$ range($Q$)}
	 		{
	 		${\boldsymbol{\hat{\rho}}}_{j,q}(m) = [\hat{\rho}_{j1},\hdots,\hat{\rho}_{jK}] \leftarrow f({\bf{x}}_{\text{t},q}(m))$\\
		    ${\cal{L}}({\boldsymbol{\theta}}, f({\bf{x}}_{\text{t},q}(m))) = \sum_{k=1}^{K} {\hat{\rho}}_{jk}$ \\ 
		    ${\boldsymbol{\eta}}_q =  {\text{sign}}\left(\nabla_{{\bf{x}}_{\text{t},q}(m)} {\cal{L}}({\boldsymbol{\theta}}, f({\bf{x}}_{\text{t},q}(m))) \right)$\\
		    ${\bf{x}}_{\text{t},q+1}(m) = {\bf{x}}_{\text{t},q}(m) + \alpha \, {\boldsymbol{\eta}}_q$\\
		    ${\bf{x}}_{\text{t},q+1}(m) \leftarrow {\text{clip}}_{[{\bf{x}}_\text{t}(m), \epsilon]} \{{\bf{x}}_{\text{t},q+1}(m)\} $
	 
	 		}
	 		$ {\bf{x}}_j^\text{adv}(m) \leftarrow {\bf{x}}_{\text{t},Q}(m)$\\
    		${\boldsymbol{\hat{\rho}}}_{j}^\text{adv}(m) = [{\hat{\rho}}_{j1}^\text{adv}, \hdots, {\hat{\rho}}_{jK}^\text{adv}] \leftarrow f({\bf{x}}_j^\text{adv}(m))$\\
\eIf{$\sum_{k=1}^{K} {\hat{\rho}}_{jk}^\mathrm{adv} >  P_\mathrm{max}$}
        {
			${\boldsymbol{\hat{\rho}}}_{j}^\text{adv}(m) \leftarrow$ infeasible power\\
				$\mathrm{ix\_list}.\mathrm{append}(m)$
		}
		{
		${\boldsymbol{\hat{\rho}}}_{j}^\text{adv}(m) \leftarrow$ feasible power	 
		}
	 }
	 ${\bf{x}}_j^\text{adv}(\mathrm{ix\_list}) $\\
	 ${\boldsymbol{\hat{\rho}}}_{j}^\text{adv}(\mathrm{ix\_list})$ 
	}
 \caption{PGDM attack}
\end{algorithm} 
\subsection{Momentum iterative fast gradient sign method (MI-FGSM)}
The momentum iterative FGSM was proposed by Dong et al. \cite{Dong_boosting_adversarial_2018} updates the adversarial example by incorporating the velocity vector along the direction of the gradient of the loss function across multiple iterations. The gradient update in the $(i+1)$-th iteration can be calculated as  
\begin{equation}
{\bf{g}}_{i+1} = \mu \, {\bf{g}}_i + \frac{\nabla_{{\bf{x}}_{\text{t},i}}{\cal{L}}({\boldsymbol{\theta}},f({\bf{x}}_{\text{t}, i}))}{||\nabla_{{\bf{x}}_{\text{t},i}}{\cal{L}}({\boldsymbol{\theta}},f({\bf{x}}_{\text{t}, i}))||_1}\,,
\label{grad_mifgsm_update}
\end{equation}
where $\mu$ and $||\cdot||_1$ denote the decay factor and the $L_1$-norm, respectively. The algorithm for cell-based MI-FGSM attack against power allocation in a massive MIMO system is similar to that of  proposed in Algorithm~$2$, except that the  
adversarial examples are computed by updating as given by 
\begin{equation}
{\bf{x}}_{\text{t},i+1} = {\bf{x}}_{\text{t},i} + \beta \, {\text{sign}}({\bf{g}}_{i+1})
\label{adv_mifgsm}
\end{equation}
until the $I$-th iteration without performing the clipping operation in each iteration. The small step size and the number of iterations are denoted by  $\beta$ and $I$, respectively. The gradient and the step size are initialized as ${\bf{g}}_0 = 0$ and $\beta = {\epsilon}/{I}$, respectively. 

\subsection{Random perturbation}
To understand the security aspects and vulnerability of DNN models for adversarial perturbations which are not actually   random perturbations, it is also important to evaluate the DNN models by constructing random perturbations having the structure similar to that of perturbations that are generated by FGSM attack. This can be formulated as follows
\begin{eqnarray}
{\boldsymbol{\eta}} &=& \epsilon \, \, \text{sign} ({\bf{w}}) \, , \nonumber \\
{\bf{x}}_\text{rnd} &=& {\bf{x}}_\text{t} + {\boldsymbol{\eta}} \, ,
\end{eqnarray}
where ${\bf{w}} \sim {\cal{N}}({\bf{0}}, {\text{\bf{I}}})$ is the random normal distribution having zero mean and identity covariance matrix. The random perturbation of input clean samples denoted by ${\bf{x}}_\text{rnd}$  implies that UE positions are perturbed in the random directions with magnitude $\epsilon$ as opposed to well-crafted directions as in FGSM attack. Since the input to the DNN is the users position, increasing the input of DNN does not necessarily increase the predicted output of the DNN.  
The random perturbations are generated with equal sign probability as $\Pr(\epsilon) = \Pr(-\epsilon) = 0.5$. 
That is, as the random perturbation follows a normal distribution, the probability that the sign perturbation is positive will be $50\%$, and negative $50\%$.

\subsection{Transferability: Black-box attack}
In Subsections IV.A--IV.E, we have described the attack model and the white-box attacks by assuming that the adversary has access to the input of the network model. In this subsection, we further   investigate  black-box attacks, where the attacker has limited or no access to the target DNN model. The white-box attacks provide a worst-case scenario while the black-box attacks provide a realistic practical setting in the sense that the attacker does not have access to the DNN model. Transferability is an important concept for black-box attacks since adversarial examples transfer between different DNN models trained independently \cite{Papernot_transferability_in_2016}. It turns out that adversarial examples computed against a model often transfer to other models and could fool them with high probability \cite{Papernot_practical_black-box_2017}. Therefore, the adversary could train a surrogate model and generate adversarial examples with the white-box approach against this model. Due to the transferability property, the target model will be vulnerable to the adversarial examples generated against the surrogate model. In this paper, for the wireless application under study, we provide below  the procedure that explains the transferability concept for a cell-based attack with $L_\infty$-norm as: 
\begin{itemize}
\item[(1)] Train a surrogate DNN model $f_{M_1}(\cdot)$ on the training dataset  ${\bf{x}} \in {\cal{X}}$ and ${\boldsymbol{\rho}}_j \in {\cal{P}}$, $j = 1, \ldots, L$, where the DNN model architecture $M_1$ is given in Table~$1$.

\item[(2)] Generate adversarial examples ${\bf{x}}_\text{adv} = {\bf{x}}_\text{t} + {\boldsymbol{\eta}} $ on model $f_{M_1}(\cdot)$ in a white-box manner using the proposed $L_\infty$-norm FGSM, PGDM, and MI-FGSM attacks presented in Algorithm~$1$, Algorithm~$2$, and (\ref{grad_mifgsm_update})--(\ref{adv_mifgsm}), respectively. 

\item[(3)] For the same task, train a victim DNN model $f_{M_2}(\cdot)$ on the training dataset  ${\bf{x}} \in {\cal{X}}$ and ${\boldsymbol{\rho}}_j \in {\cal{P}}$, $j = 1, \ldots, L$,, where the DNN model architecture $M_2$ is given in Table~$2$. This model is unknown to the adversary. 

\item[(4)] The transferability is analyzed which is critical for black-box attacks, where the victim model $f_{M_2}(\cdot)$ is not accessible. We evaluate the adversary performance by applying different attacks against the victim model using adversarial generated examples in step (2). 

\item[(5)] To   experiment further, we study the transferability property by interchanging the role of surrogate and victim models in steps (1)--(4), i.e. the surrogate and victim models become $f_{M_2}(\cdot)$ and $f_{M_1}(\cdot)$, respectively. \end{itemize}

\section{Defense technique: Adversarial training}
\paragraph{Why adversarial training}
Before delving into the details of adversarial training as a defense mechanism against adversarial attacks, we discuss some alternative solutions, and explain why we have focused on adversarial training in this paper.

There are a number of different approaches to defend against adversarial attacks like distillation \cite{Papernot_distillation_as_2016, Carlini_towards_evaluating_2017};  detection of adversarial examples \cite{Metzen_on_detecting_2017, Feinman_detecting_adversarial_2017, Carlini_adversarial_examples_2017, Tao_attacks_meet_2018,Carlini_is_ami_2019}; thermometer encoding \cite{Buckman_thermometer_encoding_2018} and pre-processing based defenses \cite{Guo_countering_adversarial_2018, Song_pixeldefend_leveraging_2018}. But it has been shown that all these methods can be defeated \cite{Athalye_obfuscated_gradients_2018, Rice_overfitting_in_2020} and adversarial training remain as a main candidate for creating a robust network \cite{Madry_towards_deep_2018}.

Aside from the general approaches for creating a robust network, discussed in previous paragraph, for the specific problem we considered one can use re-scaling. By re-scaling we mean to re-scale the output of DNN such that it meets the total power constraint. 
Then the re-scaled power coefficient of UE $k$ in cell $j$ denoted by $\hat{\rho}_{jk}^\mathrm{R,tr}$, can be defined as
\begin{eqnarray}
\label{rescale_truth}
\hat{\rho}_{jk}^\mathrm{R,tr} = {\hat{\rho}}_{jk}^\mathrm{adv}\, \frac{\sum_{k^\prime=1}^K \rho_{jk^ \prime}}{\sum_{k^\prime=1}^{K} \hat{\rho}_{jk^\prime}^\mathrm{adv}}\,, \,\, k = 1,\ldots,K\, ,
\end{eqnarray}
where $\hat{\boldsymbol{\rho}}_{j}^\mathrm{R,tr} = [\hat{\rho}_{j1}^\mathrm{R,tr}, \ldots, \hat{\rho}_{jK}^\mathrm{R,tr}]$ and $j = 1, \ldots, L$.
There are two major drawbacks with this approach. First, this is a very specific solution that just applies to our considered problem, hence cannot be used in other applications, e.g., modulation classification, while adversarial training is a 
general approach. Second, even for the specific problem considered in this paper, the re-scaling just partially compensates for the adversarial attack.
\begin{figure}[t]
	\centering
	\includegraphics[width=3.6in,height=2.3in]{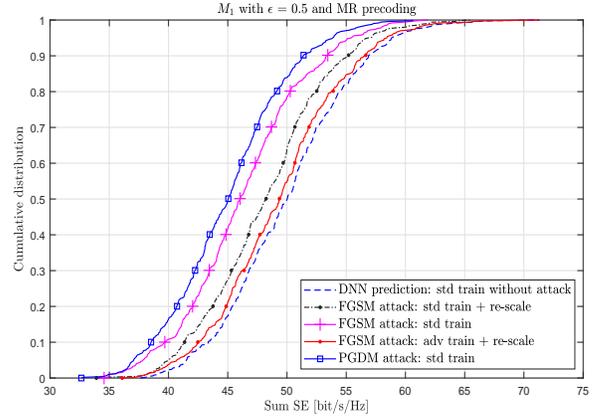}
	\caption{Cumulative distribution of sum SE for FGSM and PGDM attacks with $\epsilon = 0.5$ on $M_1$ with MR precoding.}
	\label{FGSM_PGD_minus_sum_of_ue_powers}
\end{figure}

The objective of the DNNs considered is to map users locations to power coefficients such that the product of the SINR is maximized. Another possible aim of the adversary is to minimize the sum of UEs' powers in each cell, that is, minimizing the product SINR which is approximately equivalent to minimizing the sum spectral efficiency (SE) of the multicell massive MIMO system. The downlink SE \cite{bjornson2017massive} of UE $k$ in cell $j$ is given by 
\begin{equation}
\mathrm{SE}_{jk} = \frac{\tau_d}{\tau_c}\, \mathrm{log}_2(1 + \gamma_{jk})\,, \quad [\mathrm{bit/s/Hz}]\,,
\end{equation}
where $\gamma_{jk}$ is the SINR as given in \eqref{SINR_eq}, $\tau_c$ is the coherence block, and $\tau_d$ is the number of samples used for downlink data. 

Given the aim of the adversary is to minimize the sum SE, Fig. \ref{FGSM_PGD_minus_sum_of_ue_powers} presents the cumulative distribution of sum SE
using $M_1$ with MR precoding, where the power coefficients are obtained from the DNN  with (i) standard training without any attack, (ii) standard training under adversarial attack, (iii) standard training plus re-scaling under FGSM attack, and (iv) adversarial training plus re-scaling under FGSM attack. It can be observed that when there is no adversarial attack, the sum SE $\ge 50$ [bit/s/Hz] for $50$\% of the time, whereas with FGSM attack, it is only $20$\% of the time, and with PGDM attack, it is only $18$\% of the time. Given these values, the attack has significant impact on the sum SE. 
In other words, without adversarial attack and with adversarial attack, the sum SE $\ge 50$ [bit/s/Hz] and sum SE $\leq 45$ [bit/s/Hz], respectively, is achievable for  $50$\% of the time. The loss of $\approx 5$ [bit/s/Hz] due to adversarial attack has significant impact on the achievable performance of a network. After adversarial training plus re-scaling, the loss is only  $\approx 1$ [bit/s/Hz] which has significantly improved the performance of a network. 
Furthermore, in our simulations, for the specific task considered in this paper, we did not observe any performance loss due to adversarial training on clean inputs (that is when there is no attack).
From Fig. \ref{FGSM_PGD_minus_sum_of_ue_powers} two observations can be made. First, for this specific problem, although re-scaling is possible, it just can partially  improve the performance. Second, using the adversarial training improves the performance, even when re-scaling is already applied.
This means that even for cases that re-scaling can be considered as an option, adversarial training provide further gains. Given these observations, we focus on adversarial training as a general (can be applied to any other wireless application) defense mechanism with the objective to make the DNN models robust against adversarial perturbations. 

\paragraph{Existing works and contributions} In  \cite{Madry_towards_deep_2018, Goodfellow_explaining_and_2015, Huang_learning_with_2016}, it has been shown that augmenting the training dataset with adversarial examples increases the robustness of DNNs to adversarial examples. In computer vision applications, much prior work has attempted adversarial training as a defense method against adversarial examples on   classifier datasets such as MNIST \cite{Goodfellow_explaining_and_2015}, CIFAR10 \cite{Madry_towards_deep_2018}, and ImageNet\cite{Kurakin_adversarial_machine_2017}. Furthermore, in \cite{Nguyen_adversarial_attacks_2018}, the authors  devised a regularization based defensive method against the adversarial attacks in a regression setting by inducing stability in learned functions. This method was proposed by considering the possibility that the adversarial perturbation could be caused due to instability in learned functions and the experiments were conducted on the regression datasets obtained from the UCI Machine Learning repository. The authors of \cite{Kurakin_adversarial_machine_2017} and \cite{maini_adversarial_robustness_2020} obtained a significant improvement of robustness against adversarial examples generated by the FGSM and the PGDM-based approaches, respectively. Different from the previous works, we investigate adversarial training as a defensive technique for a regression problem in the context of a wireless application. 

We can define the adversarial training as an approach in which the model learns to minimize the worst-case 
loss within the given perturbation magnitude $\epsilon$. Mathematically, this can be formulated as an optimization problem \cite{Ren_adversarial_attacks_2020} given by 
\begin{equation}
\underset{\boldsymbol{\theta}}{\text{min}}\quad
\underset{||{\boldsymbol{\eta}}||_{\infty} \leq \epsilon}{\text{max}} {\cal{L}}({\boldsymbol{\theta}},{\bf{x}}_{\text{adv}},{\boldsymbol{\rho}}_j)\,, 
\label{min_max}
\end{equation}
where ${\cal{L}}({\boldsymbol{\theta}},{\bf{x}}_{\text{adv}},{\boldsymbol{\rho}}_j)$ is the adversarial loss function \eqref{loss_function} with adversarial examples ${\bf{x}}_\text{adv} = {\bf{x}} + {\boldsymbol{\eta}}$ as the input, ${\boldsymbol{\rho}}_j$ as the original (ground-truth) output of cell $j$, and ${\boldsymbol{\theta}}$ as the model $f(\cdot)$ parameters. The worst-case loss is constrained within $L_\infty$-norm bounded perturbations of magnitude $\epsilon$.  In (\ref{min_max}), the objective of inner maximization is to determine the adversarial examples that is solved by the proposed adversarial attack methods in this paper such as FGSM or PGDM or MI-FGSM. 
The objective of the outer minimization is  to minimize the loss function by applying the standard training procedure.  
Although, the adversarial training have been extensively investigated on classification-based applications such as in computer vision and natural language processing, it is overlooked for wireless system based applications. In wireless systems, we encounter both classification-based (e.g., modulation classification \cite{Sadeghi_adversarial_attacks_2019, Kim_over_the_2020, Filipovic_adversarial_examples_2019, Flowers_evaluating_adversarial_2020, Bair_on_the_2019})
and regression-based (e.g., power allocation \cite{Sanguinetti_deep_learning_2018}) applications. In this paper, we consider the adversarial training on a more complex problem of a regression setting that is
DNN based power allocation in a multicell massive MIMO network against adversarial attacks.   

Goodfellow et al. \cite{Goodfellow_explaining_and_2015} proposed FGSM-based adversarial training to improve the robustness of a DNN by training it with both clean samples and adversarial samples crafted using FGSM. The experiments in \cite{Goodfellow_explaining_and_2015} suggested that the DNN was robust to FGSM-crafted adversarial examples, i.e., with adversarial training, the error rate on adversarial examples significantly reduced. However, the model was found to be still vulnerable to multi-step iterative-variants of adversarial attacks. In our work, we benchmark the proposed $L_\infty$-norm adversarial attacks (FGSM, PGDM, and MI-FGSM) and found that PGDM is the strongest attack having a high probability in fooling the DNN. Therefore, in our work, we consider the adversarial training with $L_\infty$-norm PGDM-crafted adversarial examples. It has also been shown in the literature that PGDM attack as a universal `first-order $L_\infty$ adversary' \cite{Madry_towards_deep_2018} and the robustness against such a first-order adversary  guarantees security across other first-order $L_\infty$-norm attacks namely FGSM \cite{Goodfellow_explaining_and_2015} and Carlini and Wagner \cite{Carlini_towards_evaluating_2017} method. This was shown by training the DNNs on the classification datasets such as MNIST and CIFAR10. To the best of our knowledge, this is the first work discussing adversarial training for creating robust DNN-based model for wireless systems.
The proposed procedure for cell-based adversarial training with $L_\infty$-norm PGDM-based adversarial examples against DNN-based power allocation in a massive MIMO system is as follows:
\begin{itemize}
\item[(1)] Train a DNN model $f_\text{std}(\cdot)$ on the training dataset ${\bf{x}} \in {\cal{X}}$ and ${\boldsymbol{\rho}}_j \in {\cal{P}}$ for each cell $j$, where $j=1, \ldots, L$. We refer to this as \textit{standard
training} since the input to the DNN is the original clean UEs positions and the output is the optimal power coefficients of UEs that are obtained using traditional optimization techniques.

\item[(2)] Generate adversarial examples using  model $f_\text{std}(\cdot)$ for each $\{{\bf{x}}(n),{\boldsymbol{\rho}}_j(n)\}$, $n=1, \ldots, N_1$ and $j=1, \ldots, L$, with the proposed $L_\infty$-norm PGDM-based adversarial attack in Algorithm~$2$.  
The generated adversarial examples dataset is denoted by $\{{\bf{x}}_\text{adv}(n),{\boldsymbol{\rho}}_j(n)\}$. 

\item[(3)] Train a DNN model $f_\text{adv}(\cdot)$ on the adversarial dataset ${\bf{x}}_\text{adv} \in {\cal{X}}$ and ${\boldsymbol{\rho}}_j \in {\cal{P}}$ for each cell $j$. We refer to this as \textit{adversarial training} since the input to the DNN is the adversarial perturbed UEs positions and the output is the ground-truth power coefficients of UEs.

\item[(4)] In \cite{Rice_overfitting_in_2020}, it has been shown that overfitting for adversarial training could result in  degraded performance for the test dataset. This implies that unlike in the standard training, overfitting in adversarially training of DNN has the property of increasing the robust test loss. Therefore, we employ \textit{validation-based early stopping} which terminates learning after the validation loss of the adversarial validation dataset does not decrease for the specified number of epochs. 

\item[(5)] In the final step, we consider the test dataset ${\bf{x}}_\text{t}$ (clean samples) different from the adversarial training and adversarial validation datasets to evaluate the robustness of the adversarially trained model $f_\text{adv}(\cdot)$ against the FGSM, PGDM, MI-FGSM, and random perturbation based attacks. 
\end{itemize}

\section{Experimental results and discussions} \label{Experimental_results}
In this section, we analyze and discuss in detail the adversarial attacks to create infeasible solution using \eqref{loss_function} and adversarial training for the given application of a regression problem in a wireless scenario.  To this end, we first investigate the vulnerability of the DNN models $M_1$ and $M_2$ built for solving power allocation problem in the downlink of a massive MIMO system with max-product SINR strategy and the precoding schemes as MR and M-MMSE using the $L_\infty$-norm of FGSM, PGDM, MI-FGSM attacks, and random perturbations. We then investigate the robustness of models  $M_1$ and $M_2$ by employing adversarial training using the adversarial generated examples from PGDM-based approach. We use the same models and the dataset that is available for the public in the given link:  \url{https://data.ieeemlc.org/Ds2Detail}. In our experiments, we consider the training and test dataset sizes as $N_1 = 329,000$ and $N_2 = 500$, respectively. For PGDM attack, we have used $\alpha = 0.01$ and $Q = 40$; for MI-FGSM attack, we have used $\mu = 0.1$, $I = 10$, and $\beta = 0.1\, \epsilon$. The system model parameters such as $K$, $L$, $M$, $P_\text{max}$, $\sigma^2$, and the cell coverage area are given in Section III. 

\subsection{Standard training: White-box attacks}
The proposed adversarial attacks in Section IV are tested to analyze and compare the effectiveness of the attack methods in terms of adversary success rate to break the DNN-based approach. The higher the value of success rate, the higher is the probability to cause erroneous predictions by DNN. In our experiments, we generate $N_2 = 500$ adversarial examples to test on the DNN  by providing these examples as an input to the DNN and analyze the trained models $M_1$ and $M_2$ to predict the output adversarial power. The aim here is to determine that the trained models with input as adversarial generated examples in a white-box manner, for how many of these examples, the DNN could provide feasible and infeasible power solutions as output. 

In all the experiments, for the test dataset $N_2 = 500$, we have validated that for clean samples (i.e., without adversarial perturbation), the DNN models $M_1$ and $M_2$ provide feasible output power solution across all the cells, i.e., $\sum_{k=1}^{K} {\hat{\rho}}_{jk} \leq P_{\mathrm{max}}$ implying that for each clean sample as input to the DNN, the predicted output of it, is always having the sum of UEs powers in each cell as less than or equal to $P_{\mathrm{max}}$, which is as expected. We apply different attacks against DL-based power allocation in a massive MIMO system, where the clean samples are the actual UEs positions. The adversarial perturbed samples are the input to the trained models $M_1$ and $M_2$, while the predicted output is tested for how many of these samples, the models predict the sum of UEs power as greater than the maximum downlink power, called as the adversary success. Since the models, $M_1$ and $M_2$ are trained under the clean input and output samples, this scenario we refer to as $\textit{standard training}$.  
 
\begin{figure}[t]
	\centering
	\includegraphics[width=3.2in,height=2.0in]{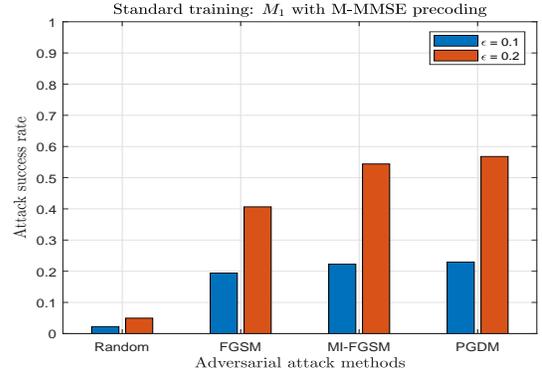}
	\caption{Standard training: Adversary success rate of different white-box attacks on $M_1$ with M-MMSE precoding.}
	\label{fig1}
\end{figure}

\begin{figure}[t]
	\centering
	\includegraphics[width=3.2in,height=2.0in]{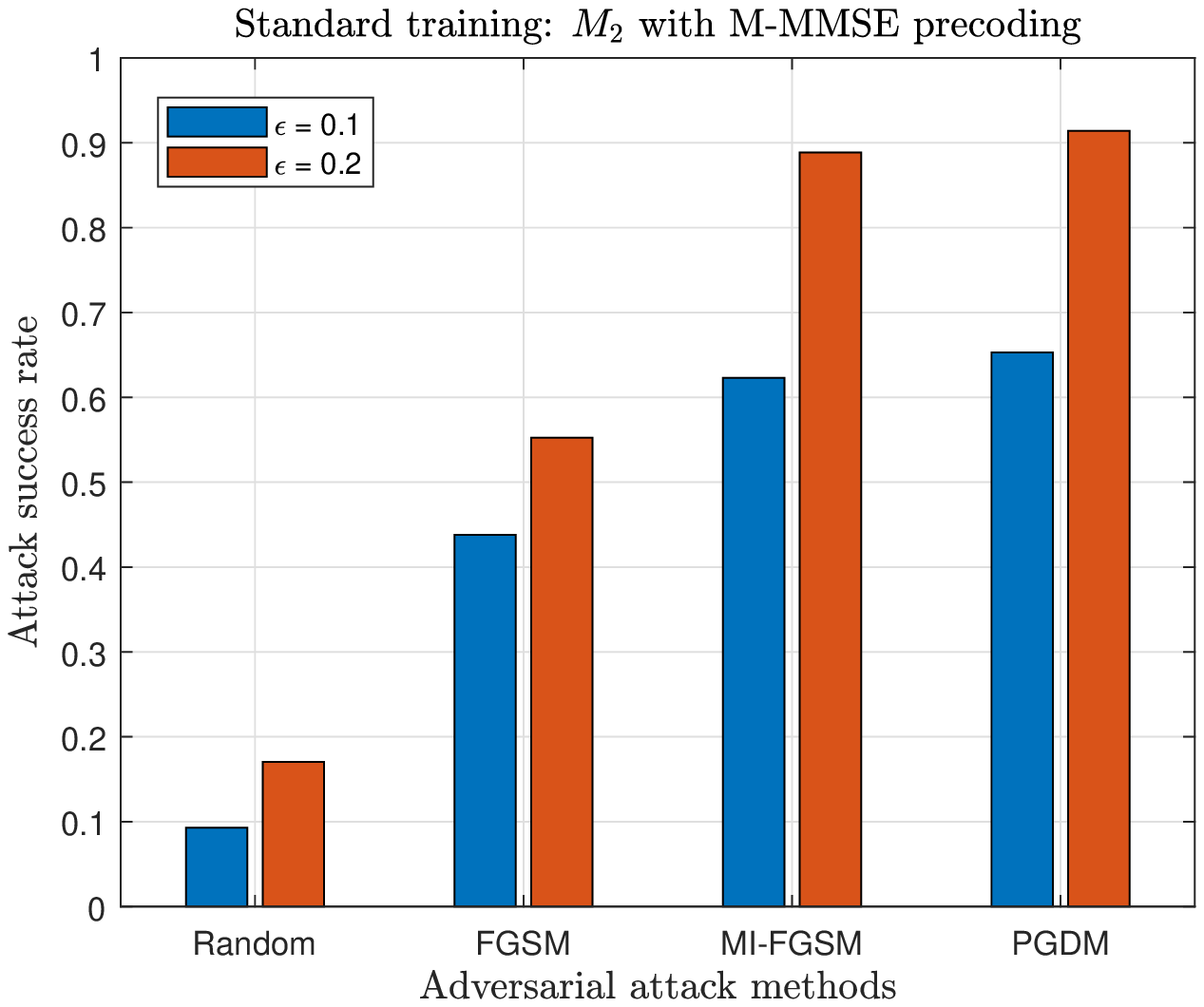}
	\caption{Standard training: Adversary success rate of different white-box attacks on $M_2$ with M-MMSE precoding.}
	\label{fig3}
\end{figure}

\begin{figure}[t]
	\centering
	\includegraphics[width=3.2in,height=2.0in]{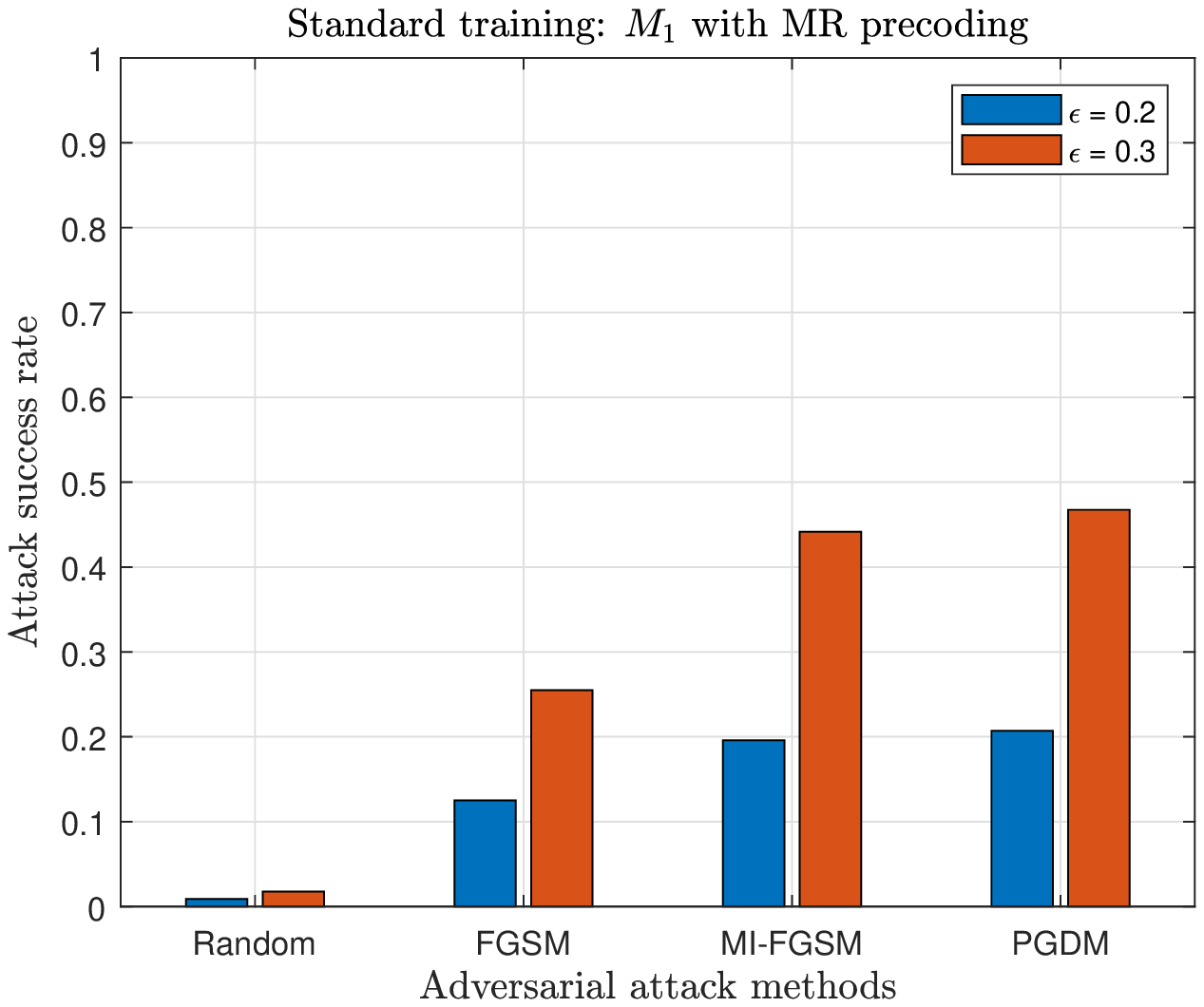}
	\caption{Standard training: Adversary success rate of different white-box attacks on $M_1$ with MR precoding.}
	\label{fig5}
\end{figure}

\begin{figure}[t]
	\centering
	\includegraphics[width=3.2in,height=2.0in]{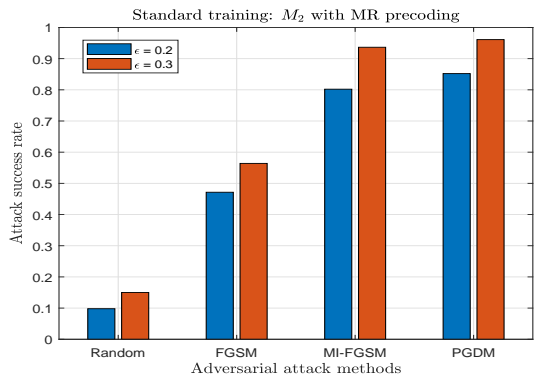}
	\caption{Standard training: Adversary success rate of different white-box attacks on $M_2$ with MR precoding.}
	\label{fig7}
\end{figure}

Figs. \ref{fig1}--\ref{fig7} present the cell-based comparison of adversary success rate of random perturbations, FGSM, MI-FGSM, and PGDM  attacks to fool the networks against $M_1$ with M-MMSE precoding, $M_2$ with M-MMSE precoding, $M_1$ with MR precoding, and $M_2$ with MR precoding, where all the models are trained with standard training. In Figs. \ref{fig1} and \ref{fig3}, the experiments are conducted with the perturbation magnitudes of $\epsilon = 0.1$ and $\epsilon = 0.2$. These $\epsilon$ values transform to the distance perturbations of locations of UEs as $d_\epsilon = 14.14$cm and $d_\epsilon = 28.28$cm. In Figs. \ref{fig5} and \ref{fig7}, we have considered $\epsilon = 0.2$ and $\epsilon = 0.3$ that corresponds to $d_\epsilon = 28.28$cm and $d_\epsilon = 42.42$cm, respectively. It has to be noted that for random perturbations, the displacement of UEs positions by $d_\epsilon$ occur in the random directions with a probability $\Pr(\epsilon) = \Pr(-\epsilon) = 0.5$, while for the gradient-based adversarial perturbation, the displacement of UEs occur in the gradient direction such that the loss is maximized. 

From the figures, it can be observed that the adversarial success rate in fooling the networks is extremely high even for a very small adversarial perturbation magnitude. 
Further, increasing the value of $\epsilon$ results in increase in the number of samples predicted as infeasible solution (i.e., increase in the adversary success rate). The adversarial success rate for the iterative-based gradient update attack (i.e., MI-FGSM and PGDM) is higher than the one-step gradient update attack (i.e., FGSM and random perturbations). 

For $M_1$ in Fig. \ref{fig1} with M-MMSE precoding and $\epsilon = 0.2$, the attack success rate for white-noise based random perturbations is less than $5\%$, while for PGDM it is $58\%$. In Fig. \ref{fig3}, for model $M_2$ with M-MMSE precoding having more trainable parameters as compared to $M_1$, thus, with $\epsilon = 0.2$, the attack success rate for random perturbations is $18\%$ and for the PGDM, it is $91\%$. For $M_1$, in Fig. \ref{fig5}, with MR precoding and $\epsilon = 0.3$, the success rate for random perturbation is less than $2\%$ and for PGDM it is $48\%$. 
For $M_2$, in Fig. \ref{fig7}, with MR precoding  and $\epsilon = 0.3$, the success rate of attacks for random perturbation is  $15\%$ and for PGDM, it is $96\%$. By comparing Figs. \ref{fig3} and  \ref{fig7}, it shows that $M_2$ with MR precoding is less vulnerable to adversarial attacks than $M_2$ with M-MMSE precoding. An intuitive reason for this could be that the M-MMSE precoding considers interference also into account which is not the case with MR precoding, therefore, with the increase in the model parameters as in $M_2$, it might be easy to attack the model learned through M-MMSE precoding.  A general remark is that a DNN architecture with less trainable parameters as $M_1$ is less vulnerable against adversarial attacks as compared to the architecture with more trainable parameters as in $M_2$, which is an intuitive result \cite{Goodfellow_explaining_and_2015}. Furthermore, MI-FGSM and PGDM algorithms were proposed to improve the performance of FGSM by conducting a iterative optimizer across multiple iterations; thus, the adversary success rate of MI-FGSM and PGDM algorithms is always higher than FGSM which is a one-step attack algorithm.

\subsection{Adversarial training}
\begin{figure}[t]
	\centering
	\includegraphics[width=3.2in,height=2.0in]{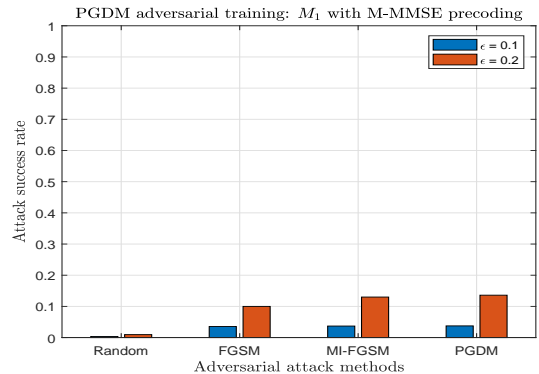}
	\caption{Adversarial training: Adversary success rate of different white-box attacks on $M_1$ with M-MMSE precoding.}
	\label{fig2}
\end{figure}
\begin{figure}[t]
	\centering
	\includegraphics[width=3.2in,height=2.0in]{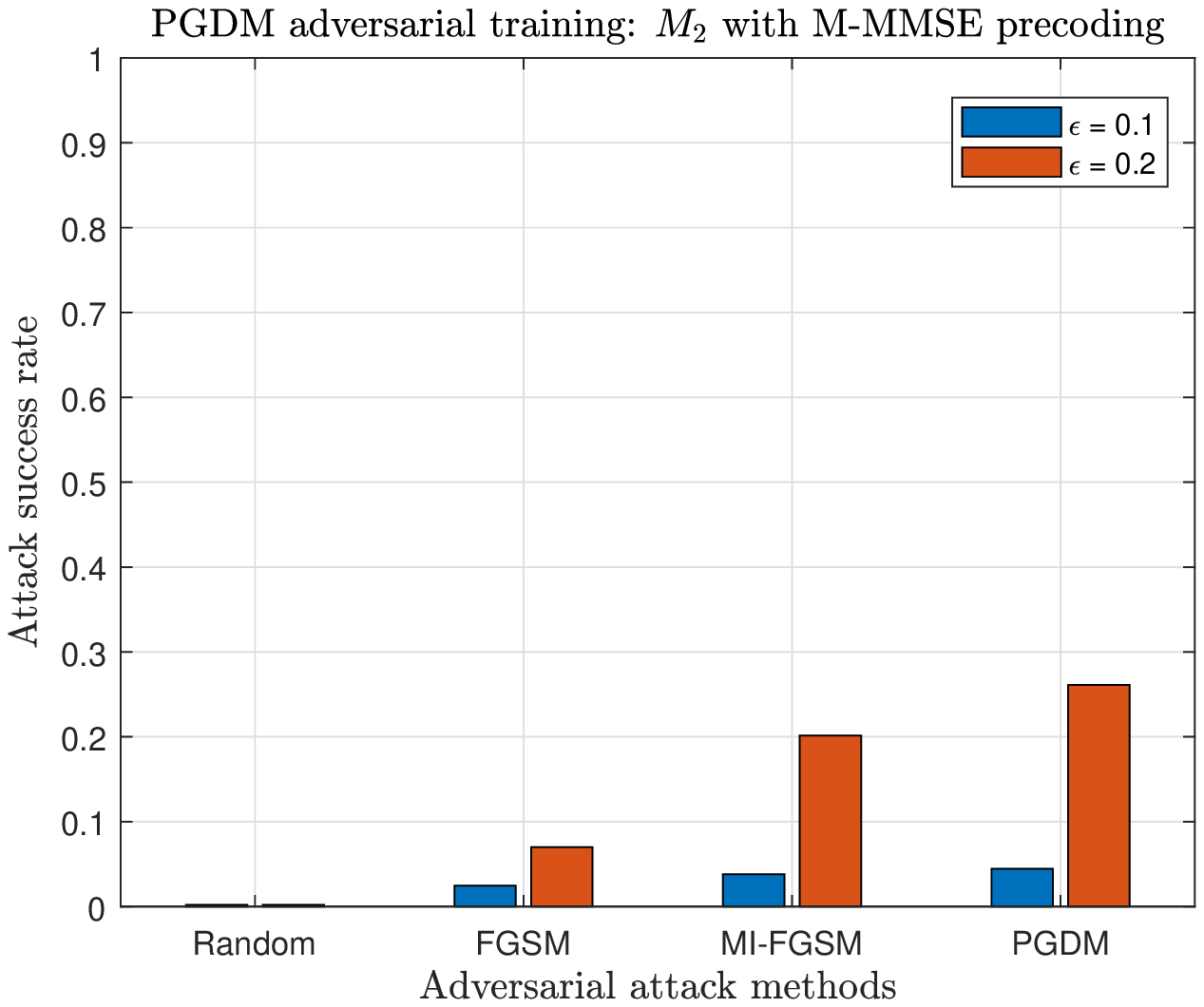}
	\caption{Adversarial training: Adversary success rate of different white-box attacks on $M_2$ with M-MMSE precoding.}
	\label{fig4}
\end{figure}
\begin{figure}[t]
	\centering
	\includegraphics[width=3.2in,height=2.0in]{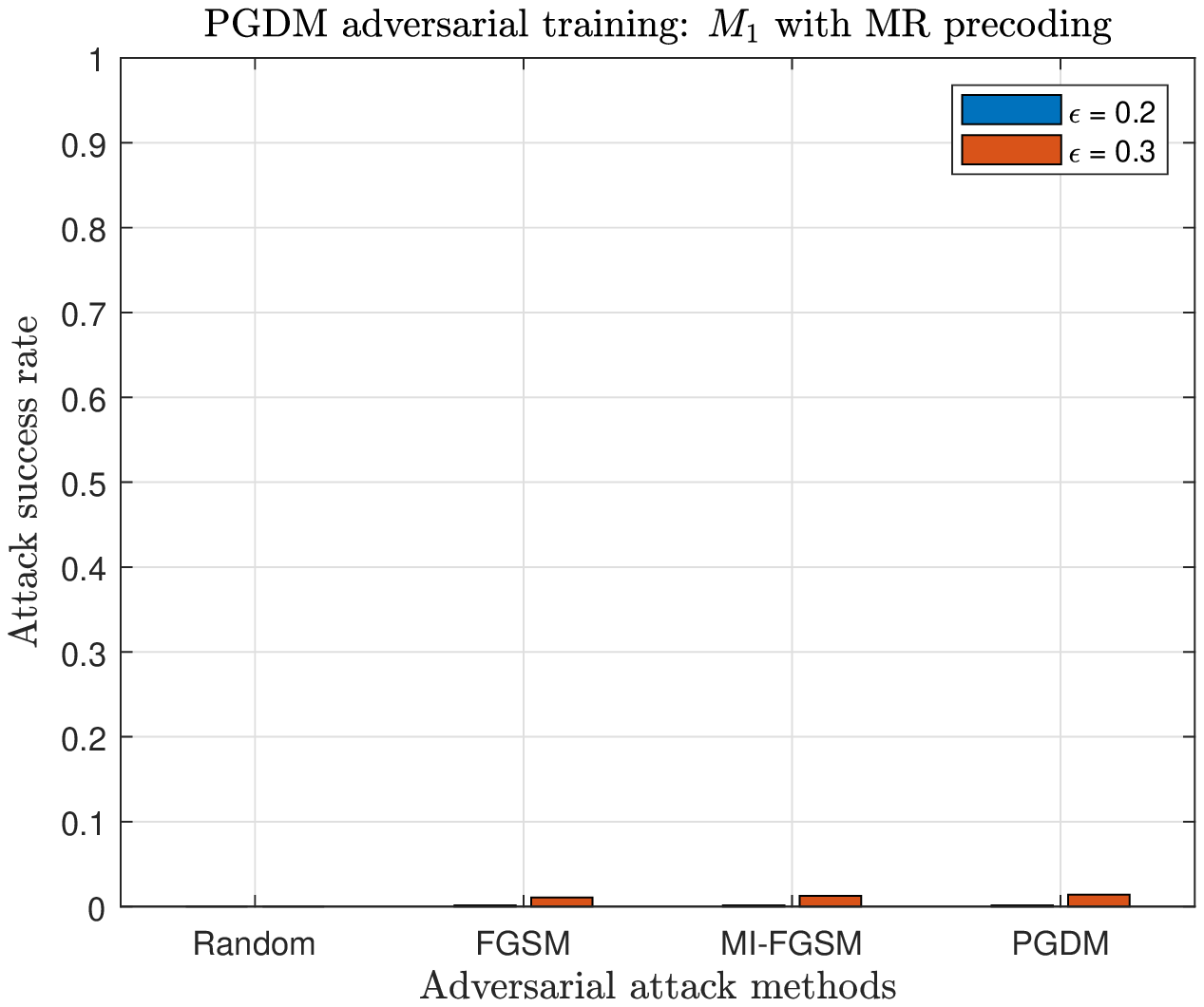}
	\caption{Adversarial training: Adversary success rate of different white-box attacks on $M_1$ with MR precoding.}
	\label{fig6}
\end{figure}
\begin{figure}[t]
	\centering
	\includegraphics[width=3.2in,height=2.0in]{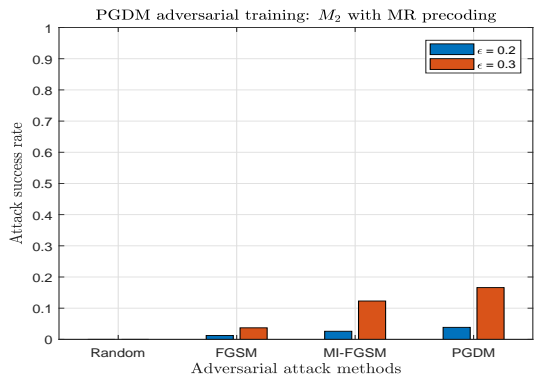}
	\caption{Adversarial training: Adversary success rate of different white-box attacks on $M_2$ with MR precoding.}
	\label{fig8}
\end{figure}
The adversarial training procedure is discussed in detail in Section V. We first generate the adversarial examples using PGDM-based approach and 
then train the DNN model $f_{\text{adv}}(\cdot)$ with these examples.  We refer to this as $\textit{adversarial training}$ since the models $M_1$ and $M_2$ are trained under the adversarial perturbed input samples and the ground-truth output samples. To test for robustness, we apply the adversarial attack methods on the adversarially trained models. Therefore, Figs. \ref{fig2}--\ref{fig8} present cell-based comparison of the attacker success rate using random perturbations, FGSM, MI-FGSM, and PGDM attacks to fool the networks against $M_1$ with M-MMSE precoding, $M_2$ with M-MMSE precoding, $M_1$ with MR precoding, and $M_2$ with MR precoding. 

A common observation for the adversarially trained models across the Figs. \ref{fig2}--\ref{fig8} is that for random perturbation, the DNN does not predict any infeasible solution, 
i.e., with random perturbations at the input of the DNN, essentially there is no occurrence of infeasible solution. 
This is an important property as in practice, we expect a certain amount of randomness at the input of the DNN, thus if the models are adversarially trained, then the DNNs are completely robust for such unwanted random perturbations.

In Fig. \ref{fig2}, for $M_1$ with $\epsilon = 0.1$, the attacker success rate on the adversarially trained model against FGSM, MI-FGSM, and PGDM attacks is less than $3\%$, which is a very good improvement in terms of robustness as compared to that of the standard trained model in Fig. \ref{fig1}. Even for $\epsilon = 0.2$, the robustness has significantly increased, where the attacks are having the success rate of $\leq 13\%$. In Fig. \ref{fig4}, we have similar observations that for $M_2$ with M-MMSE precoding after adversarial training, the model is robust against attacks. The adversary success rate against FGSM, MI-FGSM, and PGDM is under $3\%$ for $\epsilon = 0.1$, while with $\epsilon = 0.2$, for FGSM the adversary rate is $8\%$ and for iterative-based attacks, it is $20\%-25\%$. 

Fig. \ref{fig6} shows the attacks success rate on adversarial trained model $M_1$ with MR precoding for $\epsilon = 0.2$ and $0.3$. For these values of $\epsilon$, we notice that the adversarial success rate is very low ($<1.5\%$) against all the attacks, implying that the model is extremely robust. 
Fig. \ref{fig8} depicts the case of $M_2$ with MR precoding, in which we observe that for $\epsilon = 0.2$, the attack success rate is less than $3.5\%$ against all the attacks, while for $\epsilon = 0.3$, the FGSM has a success rate of $3.7\%$ and iterative-based has $12-18\%$. 
When we compare the Figs. \ref{fig1}--\ref{fig7} of standard training with the Figs. \ref{fig2}--\ref{fig8} of adversarial training, we observe that the adversarial training is more robust against attacks on the models trained with MR precoding than with the models trained with M-MMSE precoding.  

\subsection{Black-box attacks}
In this subsection, we discuss the black-box attacks using the transferability property, the detailed procedure of this is presented in Subsection IV.G. 
In other words, we investigate the attacks transferability for our regression problem in a wireless setting. We first consider the case that the adversary has full knowledge of $M_2$, where the adversary can potentially train this as a surrogate model and then generate the adversarial examples against FGSM, MI-FGSM, and PGDM attacks. To analyze the property of transferability, we assume that the adversary does not have access to the model $M_1$ and hence is the victim model. Thus, $M_1$ is analyzed for the vulnerability against the adversarial examples crafted using $M_2$. 

\begin{figure}[t]
	\centering
	\includegraphics[width=3.2in,height=2.0in]{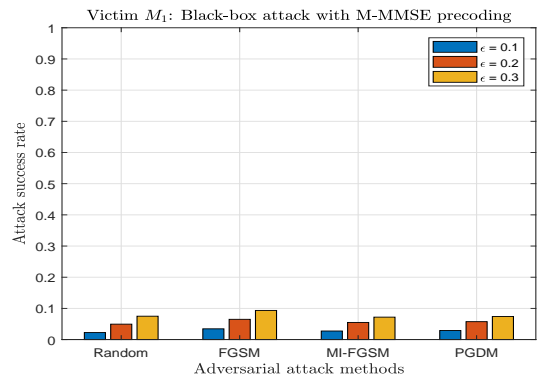}
	\caption{M-MMSE precoding: Adversary success rate of different black-box attacks with $M_1$ being victim.}
	\label{fig9}
\end{figure}

\begin{figure}[t]
	\centering
	\includegraphics[width=3.2in,height=2.0in]{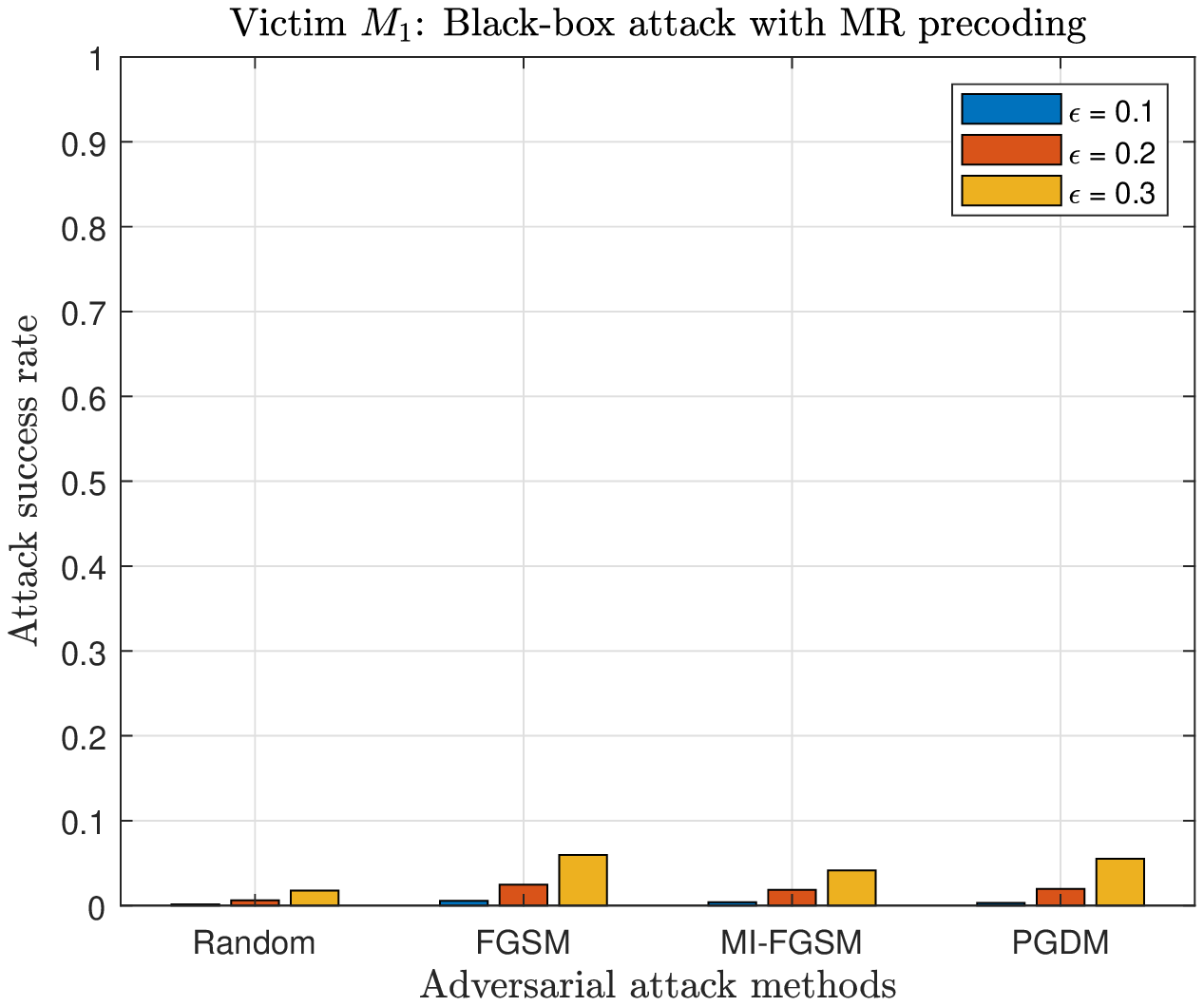}
	\caption{MR precoding: Adversary success rate of different black-box attacks with $M_1$ being victim.}
	\label{fig10}
\end{figure}

Figs. \ref{fig9} and \ref{fig10} illustrate the adversarial success rate against black-box attacks for the case of trained substitute model $M_2$ with M-MMSE and MR precoding schemes, respectively, while the victim model is $M_1$. We have shown the results for $\epsilon = 0.1, 0.2$ and $0.3$. For the victim model $M_1$, trained under standard training with both M-MMSE and MR schemes, the adversarial success rate is always below $10\%$ across different attacks. Thus, the model $M_1$ is robust for both MR and M-MMSE precoding schemes indicating poor transferability. From the figures, it can be observed that the class of gradient-based attacks to maximize the loss does not have a large impact on the DNN as compared to the random perturbations.  
\begin{figure}[t]
	\centering
	\includegraphics[width=3.2in,height=2.0in]{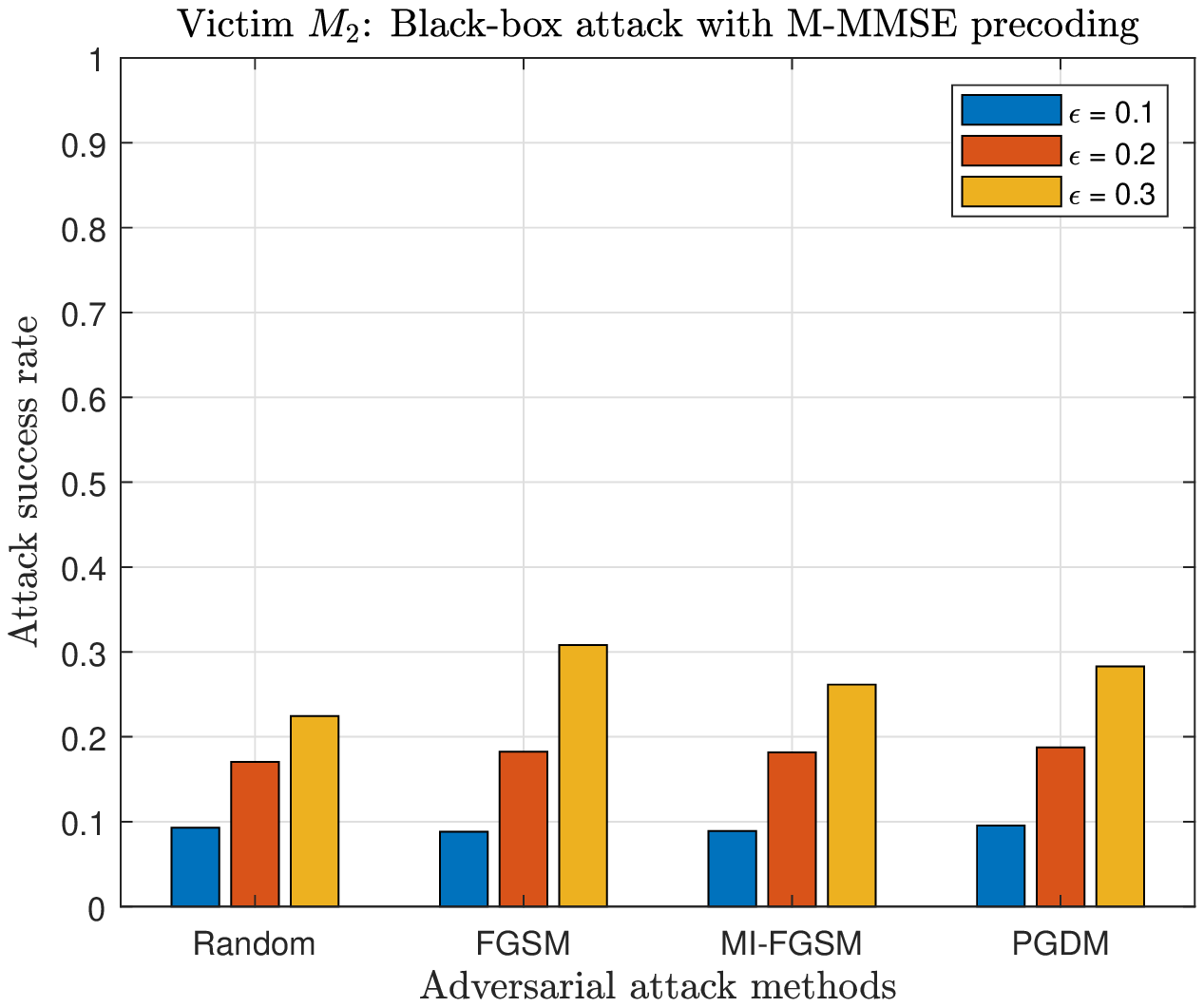}
	\caption{M-MMSE precoding: Adversary success rate of different black-box attacks with $M_2$ being victim.}
	\label{fig11}
\end{figure}
\begin{figure}[t]
	\centering
	\includegraphics[width=3.2in,height=2.0in]{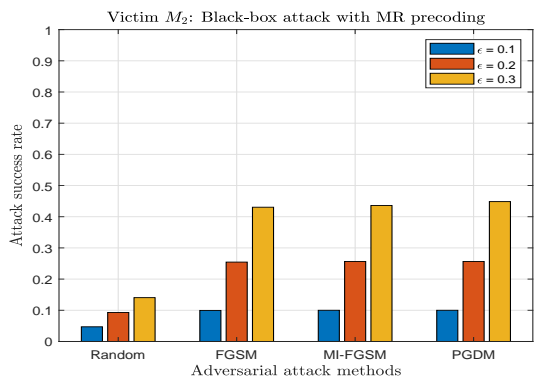}
	\caption{MR precoding: Adversary success rate of different black-box attacks with $M_2$ being victim.}
	\label{fig12}
\end{figure}

We now consider the case that adversary having the  knowledge of the substitute model $M_1$, while the victim model is $M_2$. Figs. \ref{fig11} and \ref{fig12} present the adversarial success rate in fooling the DNN victim model $M_2$, against black-box attacks for the case of trained substitute model $M_1$ with M-MMSE and MR precoding schemes, respectively. In Fig. \ref{fig11} for $\epsilon = 0.3$ with M-MMSE precoding, the success rate of FGSM attack is relatively higher than the iterative-based attacks, nevertheless, the model $M_2$ is still robust against different attacks as the success rate of adversaries is around $20-30\%$, that is way less as compared to the success rate of white-box attacks. In Fig. \ref{fig12}, for $\epsilon = 0.3$ with MR precoding, the success rate of the gradient-based attacks is around $45\%$, while for random perturbations it is less than $15\%$. The same observation holds for $\epsilon = 0.1$ and $0.2$, the success rate of adversarial attacks are the same at $10\%$ and $26\%$, respectively. The results from Fig. \ref{fig11} suggest that DNN trained with M-MMSE precoding has a poor transferability, while Fig. \ref{fig12} suggest that DNN trained with MR precoding has a good transferability. It is not obvious to generalize this claim as the obtained results are using only the models $M_1$ and $M_2$, more extensive experimentation is required to be tested on different models and by averaging over multiple runs which will be considered in the future work. One intuitive reasoning for M-MMSE being poorly transferable could be that the M-MMSE precoding is more intensive in terms of algorithm and computation as compared to the MR precoding, thus making it more difficult to learn and to apply transferability property.

\section{Conclusion}
We have presented different gradient-based adversarial attacks against the DNN-based power allocation algorithm for massive MIMO systems. For white-box attacks, we found that among the proposed attacks, PGDM is the most successful in terms of  breaking the considered  system. We then employed adversarial training by using the PGDM attack as a defense technique to improve the robustness of the DNN models against adversarial and random perturbations. We showed that the adversarial training is effective  by assuming a worst-case attack scenario to obtain a lower bound on the  performance. The discussion on power scaling is also presented and showed that it is problem specific, that does not help to create robust models against perturbations. 
We specifically observed that the model with less trainable parameters is less vulnerable irrespective of the precoding schemes. 
Furthermore, we analyzed the performance in terms of adversary success rate for black-box attacks using the transferability property for different precoding schemes. For generality of the performance analysis of the black-box attacks, we  tested on two models, $M_1$ and $M_2$ as the victim models when $M_2$ and $M_1$ as the substitute models. For all performance evaluation,  we used the publicly available dataset \cite{Sanguinetti_deep_learning_2018} providing max-product SINR power allocation strategy in a massive MIMO system with MR and M-MMSE precoding schemes.
\bibliographystyle{IEEEtran}
\bibliography{References}
\end{document}